\documentclass[a4paper,fleqn]{cas-dc}
\usepackage{hyperref}
\usepackage[numbers]{natbib}
\usepackage{subcaption}
\usepackage{graphicx}
\usepackage{float}
\usepackage{pgfplots}
\usepackage{microtype}
\usepackage{amsmath}
\usepackage{mathtools}
\usepackage[T1]{fontenc}
\usepackage{tikz}

\usepackage{xcolor}
\definecolor{DBlue}{HTML}{003A7D}
\definecolor{MBlue}{HTML}{008DFF}
\definecolor{CPink}{HTML}{FF73B6}
\definecolor{CGreen}{HTML}{4ecb8d}
\definecolor{COrange}{HTML}{FF9D3A}
\definecolor{CYellow}{HTML}{F9E858}
\definecolor{CRed}{HTML}{D83034}
\definecolor{CPurple}{HTML}{C701FF}

\usetikzlibrary{shapes,arrows,positioning,calc}

\pgfplotsset{compat=1.18}

\def\tsc#1{\csdef{#1}{\textsc{\lowercase{#1}}\xspace}}
\tsc{WGM}
\tsc{QE}
\tsc{EP}
\tsc{PMS}
\tsc{BEC}
\tsc{DE}

\begin{document}
	\let\WriteBookmarks\relax
	\def\floatpagepagefraction{1}
	\def\textpagefraction{.001}
	
	\shorttitle{Evaluating Mesh Reconstruction Methods for Crop Phenotyping}
	
	\shortauthors{Singh et~al.}
	
	\title [mode = title]{Evaluating Mesh Reconstruction Methods for Crop Phenotyping}
	
	
	%
	\author[1]{Karanvir Singh}[type=editor,
	orcid=0009-0003-0484-119X]
	
	
	\ead{karanvir.21csz0016@iitrpr.ac.in}
	
	
	\author[2]{Theo Morales}[orcid=0000-0002-2275-0895]
	\ead{tmorales@tcd.ie}
	\author[2]{Binh-Son Hua}[orcid=0000-0002-5706-8634]
	\ead{binhson.hua@tcd.ie}
	\author[1]{Mukesh Saini}[orcid=0000-0003-2215-9365]
	
	\ead{mukesh@iitrpr.ac.in}
	
	
	\affiliation[1]{organization={Indian Institute of Technology Ropar},
		city={Rupnagar},
		citysep={}, 
		postcode={140001}, 
		state={Punjab},
		country={India}}
	
	\affiliation[2]{organization={Trinity College Dublin, The University of Dublin},
		city={College Green},
		citysep={}, 
		postcode={D02 PN40}, 
		state={Dublin 2},
		country={Ireland}}
	
	
	
	\begin{abstract}
		Phenotyping an agricultural crop is crucial for studying its entire life cycle, as it provides vital insights to improve yield and, ultimately, food production.
        Doing the same for crops grown on remote sites is a challenge for the specialists who cannot be available on-site.
        3D reconstruction techniques offer a promising solution to this problem by enabling crop digitization, allowing specialists to access the resulting 3D crop models from anywhere at any time.
        In this work, we evaluate recent 3D reconstruction pipelines for crop phenotyping.
        We focus on 7 mesh reconstruction pipelines and measure the fidelity and consistency of their outputs qualitatively and quantitatively.
        Our results suggest that the meshes produced by the GGGS, PGSR, and 2DGS are preferable to the other pipelines, owing to their quantitative metrics and visually pleasing outputs.
        The GGGS pipeline is better than the second-best pipeline (2DGS) by about 27\% on the radar chart with 5 dimensions, namely, User ratings, Chamfer distance, LPIPS, PSNR, and SSIM.
	\end{abstract}
	
	

	
	\date{\today}
	\maketitle
	
	\section{Introduction}
	Food production is of utmost importance for human survival and prosperity, as it enables humankind to persevere and flourish.
	The sustainability of food production depends on crop yield, which in turn is affected by crop health.
    Phenotypes are the observable properties of a crop at different stages of its life cycle, which can help to deduce crop health.
	The process of identifying these phenotypes is called Crop phenotyping.
	Many applications, such as breeding, crop management, and crop processing~\cite{Pieruschka2019Plant}, depend on the crop phenotyping process for their success.
	
	Visual analysis is one of the foundational techniques for phenotyping, in which experts manually inspect crops for leaf color, size, and shape, flowers, or signs of disease.
    However, it is often difficult for experts to be physically available on-site to perform such inspections.
    Hence, there is a strong need to accurately digitize the crops in a format that is efficient to work with, even in remote settings.
    A 3D triangle mesh is a widely adopted format for representing and visualizing digitized scenes or objects. It also allows editing, animating, and relighting the digitized scenes~\cite{Guedon2024sugar}, which is helpful for crop phenotyping and growth models~\cite{Singh2026dtmorphing}.
    The process of obtaining a 3D mesh from multiple 2D images of a scene or an object is called 3D reconstruction.
    Traditionally, 3D reconstruction was done using traditional computer vision-based approaches.
    Recently, there has been a shift from traditional to modern deep-learning-based techniques~\cite{liu2025survey}.

    
    Despite the maturity of this field, its application in the crop phenotyping domain remains non-trivial, as the phenotyping process requires meshes to have clear photometric details to represent the crop in the real world with high fidelity.
	These details include all morphological elements, such as color, leaf count, and all parts of the crop's shoot system.
    Furthermore, these 3D meshes should also be efficient to load, render, and store for the experts working at their workstations remotely.

    Although there are studies that explore the traditional~\cite{Vazquez2016Imagingrev} and modern~\cite{yu2024sensors} 3D reconstruction pipelines from an agri-tech perspective in an exhaustive manner.
    But, to the best of our knowledge, there is no such study that experimentally compares 3D mesh reconstruction pipelines in agricultural settings.
	In this work, we provide a comparative analysis of 3D mesh reconstruction pipelines for crop phenotyping.
    On exploring the current literature for the reconstruction pipelines which output 3D triangle meshes, the following pipelines are selected for evaluation: Alicevision~\cite{Griwodz2021alicevision}, SuGaR~\cite{Guedon2024sugar}, 3DGS-to-PC~\cite{Stuart20253dgstopcmesh}, NeRF2Mesh~\cite{Tang2023nerf2mesh}, 2DGS~\cite{Huang20242DGS}, PGSR~\cite{Chen2024PGSR}, and GGGS~\cite{Zhang2026GGGS}.
    We also publicly release our \emph{Cauliflower-13} dataset, which was briefly explored in our preliminary work~\cite{Singh2024cropvr}.
	This dataset captures a growing cauliflower plant for 13 consecutive days.
    Each sample for the day consists of RGB images captured from 120 camera angles, consistently maintained using a photogrammetry setup.
	Furthermore, we perform quantitative analysis of the pipelines using 4 metrics: the Chamfer distance~\cite {Goranci2025chamferalgos}, PSNR~\cite{Fardo2016PSNR}, SSIM~\cite{Nilsson2020SSIM}, and LPIPS~\cite{Zhang2018LPIPS}.
	We also validate them qualitatively through a user study in which we ask questions about the perceived realism of leaf veins, leaf edges, stems, petioles (branches), and soil.
	We find that the GGGS pipeline is the best out of the seven, beating the second-best, 2DGS, by 27\% on the radar chart.
    
	In summary, our contributions are: 
	\begin{itemize}
		\item A new dataset for benchmarking 3D reconstruction pipelines for crop phenotyping; 
		\item An up-to-date evaluation study of traditional and state-of-the-art methods for mesh reconstruction on crop phenotyping data; 
		\item An evaluation protocol with documentation to support future research and evaluations in this domain.
	\end{itemize}


    \section{Background}
    3D reconstruction pipelines can be classified into Traditional and Deep learning-based~\cite{yu2024sensors}.
    Traditional pipelines rely on geometry and image processing techniques, which involve sub-processes such as camera calibration, camera pose estimation, feature matching, and geometric calculations before reconstructing the final 3D scene.
	Deep learning pipelines learn the mappings from input images to 3D scenes directly, without considering the above-mentioned intermediates.
    But recently, a new paradigm, Gaussian Splatting, has emerged, which is based on the intersection of traditional and deep learning paradigms.

    \subsection{Traditional Pipelines}
	Traditional pipelines can be further classified into active and passive pipelines.
	The active pipelines use real-time sensors that emit a signal to measure the position of the target surface as a collection of relative depth values, which in turn is used to reconstruct the actual target in the real world.
	Kinect~\cite{Zhang2012Kinect}, Lidar~\cite{Borkowski2024Lidar}, and Laser Scanning~\cite{Baltsavias1999airlaservsphoto} are some of the prominent traditional active pipelines.
	Conversely, passive pipelines involve capturing images via a single or multiple cameras, which are then processed by 3D computer vision techniques to reconstruct the 3D surface.
	Structure from Motion (SfM)~\cite{Rodríguez2005sfm}, Multi-View-Stereo (MVS)~\cite{Strecha2008mvs}, COLMAP~\cite{Schonberger2016colmap}, and Alicevision-Meshroom~\cite{Griwodz2021alicevision} are some of the major traditional passive pipelines.
    Active pipelines are better than passive ones in terms of accuracy and real-time outputs, but they are very expensive due to their complex hardware requirements.
	Whereas the passive pipelines do not require the emission of any signals, as a benefit, there is no interference from the environment.
    All the traditional pipelines always output the 3D scene as either a point cloud or a mesh.
	
	\subsection{Deep Learning Pipelines}
\begin{figure}
        \centering
        \includegraphics[width=0.99\linewidth]{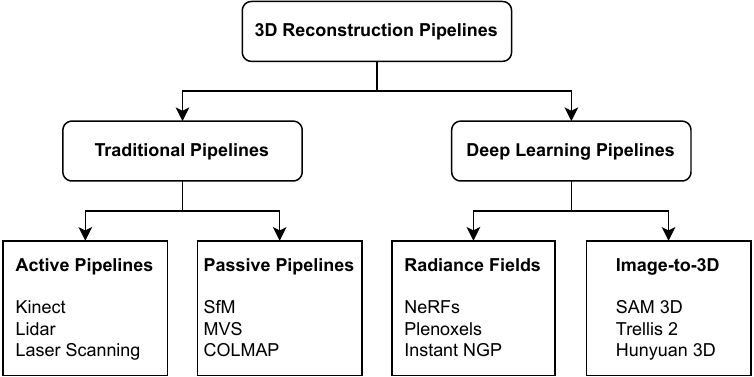}
        \caption{A basic classification of 3D Reconstruction Pipelines into Traditional and Deep learning Pipelines}
        \label{fig:traditiondeep}
    \end{figure}

    With the dawn of the deep learning era comes another 3D representation, that is, the radiance fields, notably Neural Radiance Fields (NeRF~\cite{Mildenhall2021NeRF}).
	NeRF attempts to learn a radiance field via a neural network to include the light propagation models in the output scene.
	These models capture the advanced view-dependent lighting effects such as specular highlights, reflections, and refraction, which are not possible in traditional pipelines.
	However, NeRF-based pipelines suffer from long training times and slow rendering speeds.
	Plenoxels~\cite{Keil2022plenoxels} addresses this problem by using a voxel grid instead of the neural network, resulting in two orders of magnitude faster training time but at the cost of some memory and fidelity loss.
	Instant Neural Graphics Primitives (NGP~\cite{Muller2022InstantNGP}) address the same problem by considering a smaller neural network along with a hash table.
	It attains training times even lower than Plenoxels.
	Within the same paradigm, recent foundational models called Image-to-3D have emerged, which generate a 3D mesh from a single image. 
    Sam3D~\cite{sam3dteam2025sam3d} and Trellis2~\cite{Xiang2025trellis2} are some of the notable methods in this sub-domain.
    These models are trained on massive datasets of millions of 2D images with their corresponding 3D shapes.
    An overview of both the above-mentioned paradigms is shown in Figure~\ref{fig:traditiondeep}.
	
	\subsection{3D Gaussian Splatting - A middle way.}\label{gsp}
	3D Gaussian Splatting (3DGS), discovered recently~\cite{Kerbl2023Gaussians}, is a point-based rendering method.
	It is based on elements from both of the above paradigms.
	It uses Gaussian primitives as a base for further processing, where these primitives are initialized by the traditional pipelines (SfM~\cite{Schonberger2016colmap} in the vanilla version).
	Each primitive has parameters that are optimized via deep-learning-based approaches to reduce a loss function, which is a weighted sum of two well-known terms, $L_1$ and $SSIM$~\cite{zhao2017lossfunctions}, between the rasterized and the input images (training set).
	The parameter set consists of a position in space as the mean ($\mu$) of that Gaussian primitive, and a covariance matrix ($\Sigma$), where $\Sigma$ is composed of the scale ($S$) and the rotation orientation ($R$) of the same primitive.
	There is another parameter, opacity ($\alpha$), which describes the amount of light transmission through the primitive.
	Furthermore, the color appearance when viewed from different directions is controlled by the spherical harmonics ($SH$) parameter.
	Traditional representations (point cloud, mesh) could be easily rendered via the traditional graphics pipelines as implemented in tools like Blender~\cite{blender2018blender}, Meshlab~\cite{Cignoni2008meshlab}, and Cloudcompare~\cite{cloudcompare}.
	However, for visualizing 3DGS scenes, a specialized parallel rasterization algorithm, such as SIBR Core~\cite{Bonopera2020sibr}, is used.

	\section{3D Mesh Reconstruction Pipelines}
    The above-discussed paradigms reconstruct 3D scenes as different outputs, namely point clouds, meshes, radiance fields, and splats.
    But as discussed earlier, our work considers mesh representations only, due to their widespread adoption and other application-oriented capabilities.
    This mesh preference has inspired recent NeRF and Gaussian-splatting-based works to produce mesh representations of 3D scenes rather than their usual outputs.
    After exploring the current literature for suitable works that output a mesh, we consider seven 3D reconstruction pipelines for comparison as explained below. An overview of these pipelines is presented in Table \ref{Approachestable}.
    
	
	\begin{table*}[h]
		\centering
		\caption{3D Mesh Reconstruction Pipelines Considered along with the observations made after the experiments. The preference ranks were decided from the radar chart, as in Figure \ref{fig:spiderweb}. }\label{Approachestable}\begin{tabular}{|l|l|l|l|l|l|l|}
			\hline
			\textbf{Pipeline}  & \textbf{Year}   & \textbf{Based On} & \textbf{Meshing} & \textbf{Textured} & \textbf{Preference} & \textbf{Mentionable Defects}\\ \hline
			Alicevision Meshroom~\cite{Griwodz2021alicevision}  & 2021 & Photogrammetry    & Delaunay                & Yes         &   $4^{th}$  & Poor Leaf Edges \\ \hline
			3DGS-to-PC ~\cite{Stuart20253dgstopcmesh}         & 2025 & 3DGS            & Poisson                 & No          &   $7^{th}$  & Poor Mesh \\ \hline
			SuGaR~\cite{Guedon2024sugar}               & 2024 & 3DGS              & Poisson                 & Yes & $6^{th}$    & Visible Seams  \\ \hline
			2D Gaussian Splatting~\cite{Huang20242DGS}& 2024 & 2DGS              & TSDF                    & No   &  $2^{nd}$    & Highly Smooth       \\ \hline
			NeRF2Mesh~\cite{Tang2023nerf2mesh}         & 2023   & Grid-based NeRF   & IMR                     & Yes   &   $5^{th}$   &  Soil Details missing    \\ \hline
			PGSR~\cite{Chen2024PGSR}                  & 2025 & Planar-based GS   & TSDF                    & No     &    $3^{rd}$   & High Chamfer distance  \\ \hline
			GGGS~\cite{Zhang2026GGGS}                  & 2026 & Stochastic Solids & TSDF   &    No &   $1^{st}$  & Tiny Granular Holes\\ \hline
		\end{tabular}
	\end{table*}
	
	\subsection{Alicevision Meshroom}
	Alicevision 3D reconstruction pipeline~\cite{Griwodz2021alicevision} consists of a sequential set of stages that takes the set of images as input and finally outputs a 3D mesh.
	The initial stage extracts features in all the images.
	This pipeline supports SIFT~\cite{Lowe2004sift}, DSP-SIFT~\cite{Dong2015dspsift}, and AKAZE~\cite{Alcantarilla2013akaze} features.
	Then, it proceeds with the feature matching stage, where feature descriptors are matched for all pairs of images.
	The next stage is the SfM stage, where the already found matches are fused into tracks.
	A track is considered a candidate for a 3D point if it is visible from multiple views.
	The group of these tracks is used to solve the camera calibration and generate a sparse 3D representation.
	The next stage is the depth map stage, which finds the depth of all the pixels associated with the previously calibrated cameras by Semi-Global Matching (SGM)~\cite{Hirshcmuller2008SGM}.
	The depth maps at this stage are of very low resolution, which are then upscaled with a brute force method~\cite{Griwodz2021alicevision}.
	Afterward, all the depth maps are used to re-project the depth values as 3D points, which are merged into a single dense point cloud via KD Trees.
	The next stage extracts the mesh out of the fused points via Delaunay triangulation.
	Then, the final stage is the texturing stage, in which the UV maps are extracted from the images of certain camera views that provide the best texture for the mesh region corresponding to that particular texture map.
    As a side note, while executing this pipeline in our case, we also aligned the meshes to the COLMAP sparse point cloud via the Iterative Closest Point (ICP)~\cite{Cignoni2008meshlab} to attain a comparable orientation with the other pipelines' meshes.
	
	\subsection{3DGS-to-PC}
	3DGS-to-PC~\cite{Stuart20253dgstopcmesh} pipeline converts the Gaussian splats of a scene into a very dense point cloud and ultimately into a 3D mesh.
	One of its benefits is that it doesn't need any retraining, as done in the other 3D reconstruction pipelines.
	Its input consists of the COLMAP sparse point cloud and the 3D Gaussian splats~\cite{Kerbl2023Gaussians} of the scene.
	The pipeline focuses on the sampling of points from the Gaussians, where the number of points sampled from each Gaussian is proportional to its volume.
	The Mahalanobis distance between the sampled point and the corresponding Gaussian mean is used to ensure that the outliers are not sampled.
	The color assignment for each of the newly sampled points is also a major highlight of this pipeline.
	For fast rendering, only the $0^{th}$ order spherical harmonics are used, which removes view-dependent effects of Gaussian colors.
	The idea is to balance the loss of these effects by having more newly sampled Gaussians with single color values.
	Here, the notion of contribution is introduced, which is directly proportional to the product of the opacity and transmittance values of the Gaussians.
	Each Gaussian would have a maximum contribution value, which gets updated as the training proceeds.
	The Gaussians that have this maximum contribution value lower than the overall mean contribution value are removed.
	As a result, the heavily occluded Gaussians get removed, and the Gaussians that are closer to the surface remain.
	Afterward, the point cloud is cleaned to remove any noisy outliers.
	Finally, Poisson reconstruction~\cite{Kazhdan2006poissonre} with a Poisson depth of 10 is used to extract the mesh, on which 10 iterations of Laplacian smoothing~\cite{Vollmer1999laplacesmooth} are further applied to attain the final mesh.
	
	\subsection{SuGaR}\label{sugar}
	The path from a 3DGS scene to a mesh can also be bridged by another idea of Surface-aligned Gaussian Splatting (SuGaR)~\cite{Guedon2024sugar}.
	This pipeline's primary idea is to ensure that the Gaussians output by the original 3D Gaussian Splatting pipeline are aligned to the surface.
	The input consists of a COLMAP sparse point cloud and the 3D Gaussian splats.
	There are three stages involved here, namely Regularization, Mesh Extraction, and Joint Refinement.
	
	The first stage has 3 sub-stages, where the 1st sub-stage is basically the optimization of vanilla Gaussians' properties as described in section \ref{gsp}, the 2nd sub-stage ensures that only Gaussians with opacities $\alpha$ > 0.5 remain, and the 3rd sub-stage includes new regularization term based on the signed distance function (SDF) of the surface attained by the updated Gaussians and the ideal SDF (when Gaussians lie on the surface).
	The second stage extracts a coarse mesh from the output of the first stage.
	To attain this, $\lambda$-level set points ($\lambda$=0.3) along with their normals are computed with the help of depth maps of the Gaussians along the training viewpoints.
	Then, Poisson reconstruction~\cite{Kazhdan2006poissonre} (depth 10) followed by mesh simplification via quadric error metric~\cite{Garland1997quadricmetric} is applied to reconstruct a coarse mesh from the computed points and normals.
	The third stage refines the coarse mesh by binding new Gaussians to its triangle faces.
	However, these new Gaussians have a smaller number of parameters, like 2 scaling factors instead of 3, and only 1 rotation parameter instead of 2, because they are bound to a 2D triangle, which is planar instead of the original 3D blobs.
	Finally, after further optimization, the refined textured mesh is extracted from the refined SuGaR model along with the textures of square size 8.
	
	\subsection{2D Gaussian Splatting}
	2D Gaussian Splatting (2DGS)~\cite{Huang20242DGS} is a novel idea, where the Gaussians are 2D in nature instead of 3D, implying that they are planar now.
	There are some challenges in surface (mesh) reconstruction from 3DGS because the rasterization of 3DGS lacks multi-view consistency; that is, when a 3D Gaussian is viewed from different angles, the corresponding projections are made on different intersection planes, which is not the case when the Gaussian is a 2D disk.
	Furthermore, the 3DGS representation doesn't consider surface normals, which are necessary for accurate 3D surface reconstruction.
	
	The input to the 2DGS pipeline is the sparse point cloud from COLMAP.
	This pipeline adds two new regularization terms to the usual photometric loss ($L_1$ and $SSIM$).
	One of them is about depth distortion, which ensures that the 2D Gaussians along each ray are concentrated at a specific optimized distance.
	The other term is for normal consistency, which ensures that the 2D Gaussians are locally aligned with the surface, which is done by aligning the 2D splats' normals with the normals from the depth maps.
	Apart from that, the same adaptive control strategy from 3DGS is used to increase the number of Gaussians.
	For mesh extraction, the optimized 2D splats are used to render the depth maps of the training data itself.
	Later, these depth maps are fused by Truncated Signed Distance Fusion (TSDF) using Open3D~\cite{zhou2018open3d}.
	
	\subsection{NeRF2Mesh}
	NeRF2Mesh~\cite{Tang2023nerf2mesh}, being a NeRF-based approach, extracts a textured mesh from the multi-view RGB images in a two-stage process.
	The first stage's main task is to learn the Geometry and Appearance.
	A shallow Multi-layer-perceptron (MLP) is used to learn the Geometry Density Grid.
	The Appearance is decomposed into diffuse color ($c_d$) and specular color ($c_s$).
	Diffuse color is the solid component, which is independent of any viewing directions, whereas the specular color is the shiny component, which depends upon the viewing directions.
	Both of these components are learned via two different MLPs.
	The $c_d$ being fixed can directly be converted into an RGB texture image, whereas $c_s$, being a function of viewing direction, requires a fragment shader to render it.
	The original NeRF loss function is enhanced by adding an L2 regularization term for $c_s$ along with the entropy regularization on the rendering weights.
	The Marching Cubes~\cite{Lorensen1987marchcubes} algorithm is applied on the converged Geometry Density Grid to acquire a coarse mesh, which is further refined in the second stage.
	
	In the second stage, the appearance and geometry are optimized jointly along with the vertex positions and face densities, alias iterative mesh refinement (IMR) via differential rendering~\cite{Laine2020diffrender}.
	After convergence, the geometry is refined.
	But the appearance is still in the color grid.
	Hence, the UV coordinates of the refined mesh are unwrapped.
	After which, the $c_d$ and $c_s$ are baked onto that as two different texture images $I_d$ and $I_s$.
	Additionally, during the execution of this pipeline on our dataset, the input images were down-sampled by a factor of 2 for faster convergence.
	Consequently, the final meshes were re-scaled and re-aligned to the COLMAP sparse point cloud via the ICP algorithm to attain comparable size and orientation.
	
	\subsection{PGSR}
	Planar-based Gaussian Splatting for Efficient and High-fidelity Surface Reconstruction (PGSR) focuses specifically on the depth accuracy and global geometry preservation during the reconstruction process.
	It flattens each of the 3D Gaussians along the axis for which the scaling factor is the minimum; the same axis also corresponds to the normal of the corresponding Gaussian.
	Since these flat Gaussians are fitted onto the surface, the depth maps are more consistent for the surface as compared to the cases where the Gaussians are not treated as planes, as in some previous methods (~\cite{Cheng2024GaussPropagation},~\cite{Jiang2023GaussShader}).
	
	To fit the Gaussians along the actual surface, the PGSR method optimizes the Gaussians using the loss function, which considers the terms related to flattening, photometry, and geometry.
	The flattening loss is the $L_1$ norm of the minimum scaling factors for all Gaussians.
	The photometric loss consists of the usual $L_1$ loss and $L_{SSIM}$ on the exposure-adjusted rendered images.
	The geometry loss is further composed of three aspects, namely single-view regularization, multi-view geometric consistency, and multi-view photometric consistency.
	The single-view regularization terms ensure that the final alpha-blended normal map is similar to the normal map attained via the rendered depth map.
	The multi-view geometric consistency loss considers the Homography matrix, which maps the pixels in one viewpoint to the pixels in another viewpoint (neighboring frame).
	It ensures that the forward and backward projection errors are minimized.
	The multi-view photometric consistency ensures that the normalized cross-correlation~\cite{Yoo2009NCC} between the original frame and the neighboring frame is close to $1$.
	Apart from the loss function, it follows the original 3DGS work for the initialization, along with the densification strategy as in AbsGS~\cite{Ye2024AbsGSDensi}.
	It extracts the TSDF field ~\cite{Newcombe2011TSDF} and then performs Marching Cubes~\cite{Lorensen1987marchcubes} over the TSDF field to extract the final mesh.
	
	\subsection{GGGS}
	Geometry Grounded Gaussian Splatting (GGGS) considers the Gaussians as stochastic solids whose transmission function is smooth as opposed to the step-wise nature in the case of standard 2D splats, which results in a more accurate geometry reconstruction.
	Treating them as stochastic solids allows to attain attenuation coefficients inside the Gaussian primitives, which ultimately allows for smooth optimization and accurate depth maps.
	This treatment is equivalent to the rasterization rendering of the original Gaussians.
	
	The optimization loss function is composed of the usual photometric loss~\cite{Kerbl2023Gaussians}, 2DGS consistency loss~\cite{Huang20242DGS}, and multi-view regularization loss from the PGSR~\cite{Chen2024PGSR} as mentioned previously.
	For geometric regularization, the median depth is considered, which is the point on the ray where the transmission value reaches 0.5.
	This transmission is the product of individual transmission values of the Gaussians along the ray calculated via RaDe-GS~\cite{Zhang2024RadeGS}.
	The densification strategy from the Gaussian opacity fields~\cite{Yu2024GOF} is used.
	The 3D mesh is extracted via the TSDF fusion as implemented by Open3D~\cite{zhou2018open3d}.
	
	\section{Evaluation}
	We evaluated the pipelines' outputs quantitatively and qualitatively. This section provides details of the dataset, metrics used, quantitative evaluation, user study, and result analysis.

    \subsection{Dataset}\label{datse}
	
	
	
	
	Our dataset (`\textit{Cauliflower-13}') consists of a cauliflower crop in a pot as the subject of image acquisition.
	Images were captured on a day-to-day basis from multiple views.
	The apparatus included a turntable, a tripod (Amazon Basics 60 Inch), a smartphone camera (Redmi Note 8 Pro), and a white cloth for the background.
	We captured 120 images for each day, up to a total of 13 days.
	Basically, we rotated the turntable 5 times at different height levels.
	At each height, 24 images were taken at a difference of 15$^{\circ}$ between each consecutive image at the same height.
	These points, from where the images were to be taken, were marked on the turntable.
	The heights were adjusted with the help of the rotatable knob of the tripod, which was always kept at a fixed location marked on the floor.
	The consistency of image acquisition was maintained throughout all 13 days.
	All the above-mentioned systematic specifications make this dataset a suitable testbed to evaluate the latest 3D reconstruction methods.
	
	We are releasing our dataset publicly this time, as it was explored briefly in our previous preliminary work~\cite{Singh2024cropvr}, where only the Alicevision~\cite{Griwodz2021alicevision} pipeline was used to create crop assets and visualize them in a Virtual Reality (VR) application.
    
	\subsection{Metrics}
    We used 4 metrics for evaluation, namely Chamfer distance, PSNR, SSIM, and LPIPS. The ground truth for Chamfer distance was computed via the COLMAP dense reconstruction~\cite{Schonberger2016colmap}, whereas for the other 3 metrics, the original images along with their pose information were treated as the ground truth. We discuss details of these metrics below.
    
	\paragraph{Chamfer Distance.}
	Chamfer distance is defined between two sets of point clouds as a measure of their similarity.
	It is not sensitive to the density of the point cloud unless there are very strong outliers, as it is an averaged metric.
	Chamfer Distance (CD) can be calculated as in Equation~\ref{cd}, where $\mathbf{a}$ and $\mathbf{b}$ are the vertices in the point clouds $\mathcal{A}$ and $\mathcal{B}$, respectively.
	\begin{equation}\label{cd}
		CD = \frac{1}{|\mathcal{A}|}\sum_{\mathbf{a}\in \mathcal{A}}\min_{\mathbf{b}\in \mathcal{B}}||\mathbf{a}-\mathbf{b}||^2_2 + \frac{1}{|\mathcal{B}|}\sum_{\mathbf{b}\in \mathcal{B}}\min_{\mathbf{a}\in \mathcal{A}}||\mathbf{b}-\mathbf{a}||^2_2
	\end{equation}
	
	\paragraph{PSNR.}
	Peak Signal-to-Noise Ratio measures the ratio of the square of the maximum pixel value ($MAX_{I}$) to that of the noise intensity, which is calculated as the Mean Squared Error ($MSE$) between the individual pixels of the reconstructed image and the ground truth image corresponding to the same pose at log scale in decibels (dB).
	Then, this $PSNR$ value is averaged over all the poses for a sample mesh to attain a final average PSNR metric. 
	\begin{equation}
		PSNR = 10 * log_{10}\frac{MAX_{I}^2}{MSE}
	\end{equation}
	
	\paragraph{SSIM.}
	Structural Similarity Index considers the inter-pixel dependencies, especially for the case of spatially proximate pixels, which carry the information about the structure of the object in the image~\cite{Wang2004SSIM}.
	It is calculated between two blocks ($x,y$) of the images as in Equation~\ref{ssimeq}.
	\begin{equation}\label{ssimeq}
		SSIM(x,y) = \frac{(2\mu_x\mu_y + c_1)(2\sigma_{xy}+c_2)}{(\mu_x^2+\mu_y^2+c_1)(\sigma_x^2+\sigma_y^2 +c_2)}
	\end{equation}
	where $\mu$ and $\sigma$ are the sample mean and sample variance for the corresponding $x$ and $y$ blocks, and $c_1, c_2$ are the constants determined by the dynamic range of the pixel values of the blocks.
	Then, this SSIM index is averaged over all the blocks considered in the images to attain the SSIM for the rendered and the ground truth image for a particular pose.
	Finally, the same is averaged over all the poses to attain the SSIM for the mesh.
	
	\paragraph{LPIPS.}
	Learned Perceptual Image Patch Similarity measures the similarity between two images, which coincides with human judgment~\cite{Zhang2018LPIPS}.
	It involves the use of pre-trained backbone networks to extract features from the images to be compared, which are then further used to compute the perceptual distance.
	We used the AlexNet~\cite{Krizhevsky2017Alexnet} as the backbone.
	The images to be compared also go through pre-processing to convert them to tensors of the specific shape as desired by the AlexNet backbone network.
	The LPIPS metric is calculated between the rendered image and the ground-truth images for each pose.
	It is basically the weighted difference of the extracted feature maps, where the weights are specific to the backbone network.
	Finally, the metric is averaged over all the poses to attain the final LPIPS for the mesh.
	
	\subsection{Quantitative Evaluation}
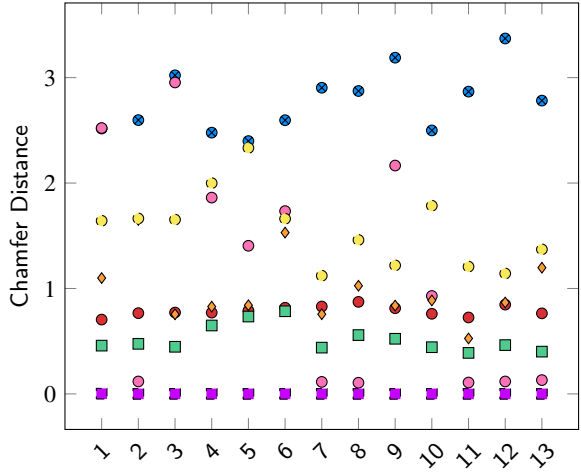
\begin{figure}
		\centering
		\begin{tikzpicture}
			\begin{axis}[width=\columnwidth,xmin=0,xmax=14,xtick={1,2,3,4,5,6,7,8,9,10,11,12,13},xticklabels={1,2,3,4,5,6,7,8,9,10,11,12,13},xticklabel style={rotate=45},ylabel={Chamfer Distance}]
				\addplot+[only marks, mark options={draw=black,fill=CRed},color=CRed]table[x=GPA,y=3dgstopc,col sep=comma]{plots/Chamfer.csv};
				\label{plot1}
				\addplot+[only marks,mark options={draw=black,fill=CGreen},color=CGreen]table[x=GPA,y=2dgs,col sep=comma]{plots/Chamfer.csv};
				\label{plot2}
				\addplot+[only marks,mark options={draw=black,fill=MBlue},color=MBlue]table[x=GPA,y=sugar,col sep=comma]{plots/Chamfer.csv};
				\label{plot3}
				\addplot+[only marks,mark=*,mark options={draw=black,fill=CPink},color=CPink]table[x=GPA,y=n2m,col sep=comma]{plots/Chamfer.csv};
				\label{plot4}
				\addplot+[only marks,mark options={draw=black,fill=COrange},color=COrange]table[x=GPA,y=pgsr,col sep=comma]{plots/Chamfer.csv};
				\label{plot5}
				\addplot+[only marks,mark options={draw=black,fill=CYellow},color=CYellow]table[x=GPA,y=gggs,col sep=comma]{plots/Chamfer.csv};
				\label{plot6}
				\addplot+[only marks,mark options={draw=black,fill=CPurple},color=CPurple]table[x=GPA,y=alicevision,col sep=comma]{plots/Chamfer.csv};
				\label{plot7}
			\end{axis}
		\end{tikzpicture}
		\caption{Chamfer distance for our Cauliflower-13 dataset \ref{plot1} 3DGS-to-PC \ref{plot2} 2DGS \ref{plot3} SuGaR \ref{plot4} NeRF2Mesh \ref{plot5} PGSR \ref{plot6} GGGS \ref{plot7} Alicevision.
			The Chamfer distance in the case of the Alicevision pipeline was the lowest owing to the ICP alignment step that was performed after the dense reconstruction stage to make it comparable to all other COLMAP-based pipelines. Besides, the Chamfer distance of the 2DGS pipeline was lower than all the pipelines except the NeRF2Mesh pipeline for some samples.}
		\label{fig:cdcauli13}
	\end{figure}

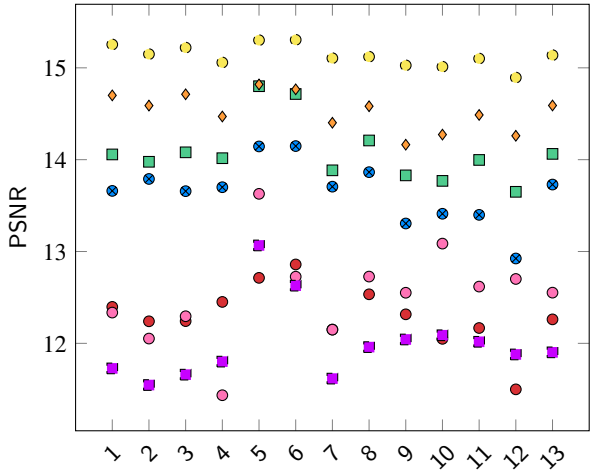
\begin{figure}
		\centering
		\begin{tikzpicture}
			\begin{axis}[width=\columnwidth,xmin=0,xmax=14,xtick={1,2,3,4,5,6,7,8,9,10,11,12,13},xticklabels={1,2,3,4,5,6,7,8,9,10,11,12,13},xticklabel style={rotate=45},ylabel={PSNR}]
				\addplot+[only marks,mark options={draw=black,fill=CRed},color=CRed]table[x=GPA,y=3dgstopc,col sep=comma]{plots/PSNR.csv};
				\label{pplot1}
				\addplot+[only marks,mark options={draw=black,fill=CGreen},color=CGreen]table[x=GPA,y=2dgs,col sep=comma]{plots/PSNR.csv};
				\label{pplot2}
				\addplot+[only marks,mark options={draw=black,fill=MBlue},color=MBlue]table[x=GPA,y=sugar,col sep=comma]{plots/PSNR.csv};
				\label{pplot3}
				\addplot+[only marks,mark=*,mark options={draw=black,fill=CPink},color=CPink]table[x=GPA,y=n2m,col sep=comma]{plots/PSNR.csv};
				\label{pplot4}
				\addplot+[only marks,mark options={draw=black,fill=COrange},color=COrange]table[x=GPA,y=pgsr,col sep=comma]{plots/PSNR.csv};
				\label{pplot5}
				\addplot+[only marks,mark options={draw=black,fill=CYellow},color=CYellow]table[x=GPA,y=gggs,col sep=comma]{plots/PSNR.csv};
				\label{pplot6}
				\addplot+[only marks,mark options={draw=black,fill=CPurple},color=CPurple]table[x=GPA,y=alicevision,col sep=comma]{plots/PSNR.csv};
				\label{pplot7}
			\end{axis}
		\end{tikzpicture}
		\caption{PSNR values for our Cauliflower-13 dataset \ref{pplot1} 3DGS-to-PC \ref{pplot2} 2DGS \ref{pplot3} SuGaR \ref{pplot4} NeRF2Mesh \ref{pplot5} PGSR \ref{pplot6} GGGS \ref{pplot7} Alicevision.
			The computed PSNR values were low for all the meshes, considering that the lighting conditions used for rendering the meshes were not exactly similar to those of the real scenes of the photographs. However, from a comparison perspective, the GGGS pipeline has the highest PSNR values, followed by PGSR, 2DGS, and SuGaR.}
		\label{fig:psnrcauli13}
	\end{figure}

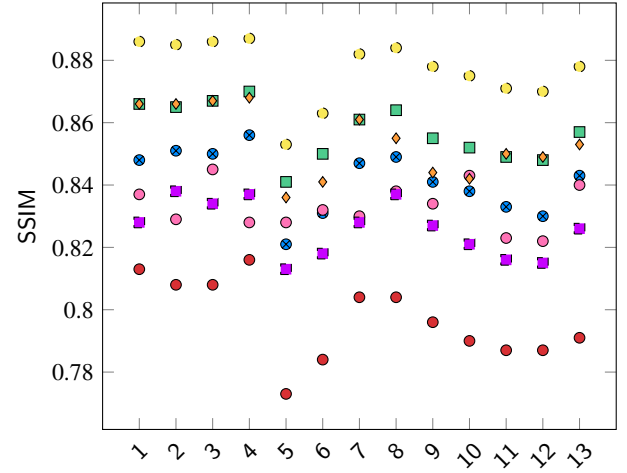
\begin{figure}
		\centering
		\begin{tikzpicture}
			\begin{axis}[width=\columnwidth,xmin=0,xmax=14,xtick={1,2,3,4,5,6,7,8,9,10,11,12,13},xticklabels={1,2,3,4,5,6,7,8,9,10,11,12,13},xticklabel style={rotate=45},ylabel={SSIM}]
				\addplot+[only marks,mark options={draw=black,fill=CRed},color=CRed]table[x=GPA,y=3dgstopc,col sep=comma]{plots/SSIM.csv};
				\label{splot1}
				\addplot+[only marks,mark options={draw=black,fill=CGreen},color=CGreen]table[x=GPA,y=2dgs,col sep=comma]{plots/SSIM.csv};
				\label{splot2}
				\addplot+[only marks,mark options={draw=black,fill=MBlue},color=MBlue]table[x=GPA,y=sugar,col sep=comma]{plots/SSIM.csv};
				\label{splot3}
				\addplot+[only marks,mark=*,mark options={draw=black,fill=CPink},color=CPink]table[x=GPA,y=n2m,col sep=comma]{plots/SSIM.csv};
				\label{splot4}
				\addplot+[only marks,mark options={draw=black,fill=COrange},color=COrange]table[x=GPA,y=pgsr,col sep=comma]{plots/SSIM.csv};
				\label{splot5}
				\addplot+[only marks,mark options={draw=black,fill=CYellow},color=CYellow]table[x=GPA,y=gggs,col sep=comma]{plots/SSIM.csv};
				\label{splot6}
				\addplot+[only marks,mark options={draw=black,fill=CPurple},color=CPurple]table[x=GPA,y=alicevision,col sep=comma]{plots/SSIM.csv};
				\label{splot7}
			\end{axis}
		\end{tikzpicture}
		\caption{SSIM metric for our Cauliflower-13 dataset \ref{splot1} 3DGS-to-PC \ref{splot2} 2DGS \ref{splot3} SuGaR \ref{splot4} NeRF2Mesh \ref{splot5} PGSR \ref{splot6} GGGS \ref{splot7} Alicevision.
			The SSIM values for the GGGS pipeline were the highest, followed by 2DGS, PGSR, and SuGaR. The 3DGS-to-PC pipeline shows the lowest SSIM values, indicating poor reconstruction based on the structural aspect.}
		\label{fig:ssimcauli13}
	\end{figure}

\begin{figure}
		\centering
		\begin{tikzpicture}
			\begin{axis}[width=\columnwidth,xmin=0,xmax=14,xtick={1,2,3,4,5,6,7,8,9,10,11,12,13},xticklabels={1,2,3,4,5,6,7,8,9,10,11,12,13},xticklabel style={rotate=45},ylabel={LPIPS}]
				\addplot+[only marks,mark options={draw=black,fill=CRed},color=CRed]table[x=GPA,y=3dgstopc,col sep=comma]{plots/LPIPS.csv};
				\label{lplot1}
				\addplot+[only marks,mark options={draw=black,fill=CGreen},color=CGreen]table[x=GPA,y=2dgs,col sep=comma]{plots/LPIPS.csv};
				\label{lplot2}
				\addplot+[only marks,mark options={draw=black,fill=MBlue},color=MBlue]table[x=GPA,y=sugar,col sep=comma]{plots/LPIPS.csv};
				\label{lplot3}
				\addplot+[only marks,mark=*,mark options={draw=black,fill=CPink},color=CPink]table[x=GPA,y=n2m,col sep=comma]{plots/LPIPS.csv};
				\label{lplot4}
				\addplot+[only marks,mark options={draw=black,fill=COrange},color=COrange]table[x=GPA,y=pgsr,col sep=comma]{plots/LPIPS.csv};
				\label{lplot5}
				\addplot+[only marks,mark options={draw=black,fill=CYellow},color=CYellow]table[x=GPA,y=gggs,col sep=comma]{plots/LPIPS.csv};
				\label{lplot6}
				\addplot+[only marks,mark options={draw=black,fill=CPurple},color=CPurple]table[x=GPA,y=alicevision,col sep=comma]{plots/LPIPS.csv};
				\label{lplot7}
			\end{axis}
		\end{tikzpicture}
		\caption{LPIPS for our Cauliflower-13 dataset \ref{lplot1} 3DGS-to-PC \ref{lplot2} 2DGS \ref{lplot3} SuGaR \ref{lplot4} NeRF2Mesh \ref{lplot5} PGSR \ref{lplot6} GGGS \ref{lplot7} Alicevision.
			A lower LPIPS score implies better reconstruction. Here, LPIPS corresponding to GGGS were the lowest, implying the best, followed by Alicevision, which was followed by the and PGSR. A lower LPIPS value is considered closer to human perception.}
		\label{fig:lpipscauli13}
	\end{figure}
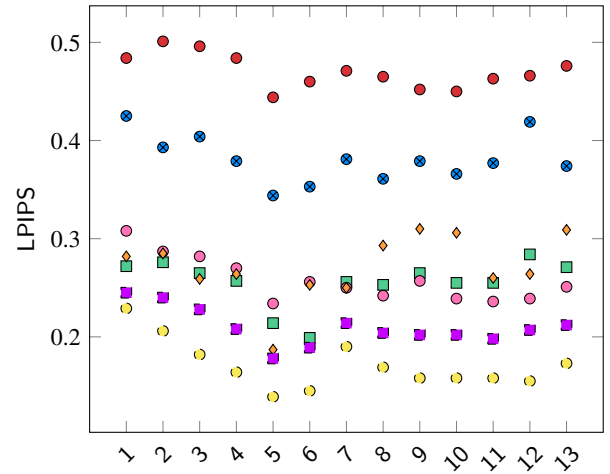

	All seven pipelines were fed the same 13 image sets for the corresponding days, yielding 13 meshes for each pipeline.
    Since the Chamfer distance calculations are concerned with point clouds, the point cloud format, which stores only the vertices of the mesh, was considered against the ground truth generated by dense COLMAP reconstruction.
	The values corresponding to this metric can be observed in Figure \ref{fig:cdcauli13}.
	For the calculations of the other three metrics, the final 3D meshes were rendered from the same poses corresponding to those of the original input images, with the ambient lighting of white color and an intensity value of 1.
	The PSNR, SSIM, and LPIPS values for the pipelines can be observed in Figures \ref{fig:psnrcauli13}, \ref{fig:ssimcauli13}, and \ref{fig:lpipscauli13}, respectively.
	Lower LPIPS values are better, implying that the final output and ground truth are more similar perceptually.
	
	\subsection{User Study}
\begin{figure}
		\centering
		\begin{tikzpicture}
			\begin{axis}[ybar,width=\columnwidth,
				bar width = 0.08\columnwidth,xtick={1,2,3,4,5,6,7},xticklabels={2DGS,Alicevision,PGSR,GGGS,NeRF2Mesh,SuGaR,3DGStoPC},,xticklabel style={rotate=45},ylabel={Final Rank from User Study}]
				\addplot+[mark options={draw=black,fill=red},color=red]table[x=GPA,y=FinalRank
				,col sep=comma]{plots/Userstudydata.csv};
			\end{axis}
		\end{tikzpicture}
		\caption{The final User Study rank was calculated by considering the 5 ranking-based questions. The overall sentiment is highest for GGGS, followed by PGSR, 2DGS, and Alicevision.}
		\label{fig:userstudyrank}
	\end{figure}
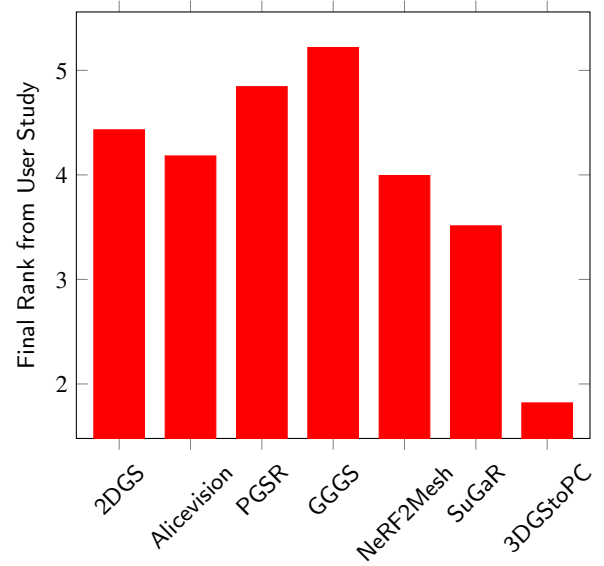

	We conducted a user study on 32 participants with basic agriculture knowledge. 
	They were asked the following 6 questions, which required ranking the meshes generated from the seven considered 3D reconstruction methods. The meshes were displayed in a web browser with interactive mouse controls to view the meshes from any desired angle. The user could swap the meshes back and forth and decide the rank they want to give for each of the questions below.
	\begin{enumerate}
		\item Please rank the leaf colors of all the methods in a relative manner from the lowest to the highest. Here, the highest implies the closest to actual images.
		\item Please rank the visibility of veins in all the methods in a relative manner from the lowest to the highest. Here, highest implies the veins are visible in the best possible manner.
		\item For which of the methods do you see artifacts (noise or defects) at the edges of the leaves? (Yes / No)
		\item Please rank the appearance of the soil details in all the methods in a relative manner from the lowest to the highest. (It may include features like the fallen leaves, pebbles, etcetera). Most details imply the highest rank.
		\item Please rank the perceived realism for the shape and appearance of the stem in all the methods in a relative manner from the lowest to the highest. Highest implies maximum realism.
		\item Please rank the perceived realism for the shape and appearance of the branches (petioles) in all the methods in a relative manner from the lowest to the highest. Highest implies maximum realism.
	\end{enumerate}
	The responses were converted into numerical data based on their rankings by mapping the ranks to an integer sentiment.
	Higher rank implies more positive sentiment towards that method.
	The highest rank had a value of 7, and the lowest rank had a value of 1.
	All the other ranking values lie between 1 and 7.
	For each ranking-related question, each mesh is given a final numerical sentiment ($Sentiment_{Mesh}$), which is calculated as
	\begin{equation}
		Sentiment_{Mesh} = \sum_{\forall i} \frac{Rank_i}{\# Participants}
	\end{equation}
	where $Rank_i$ is the numerical value for the rank assigned by the $i_{th}$ participant.
	For the question related to the artifact on the edges, the answers were Yes or No.
    The majority were naysayers for GGGS, NeRF2Mesh, and SuGaR; that is, they didn't find the artifacts on the edge of leaves, whereas for other pipelines, the majority of the participants found the artifacts on the leaf edges.
	Besides, the results of the ranking-based questions of the User study can be observed in Figure \ref{fig:UserstudyBars}.
	The final User Study Rank was calculated by averaging the ranking aspects for all 7 methods, as can be observed in Figure \ref{fig:userstudyrank}.
	
	\begin{figure*}
		\centering
		\begin{tikzpicture}
			\begin{axis}[width=\textwidth, height=0.5\textwidth,
				legend pos=north east,
				ybar,
				bar width=0.15,
				xtick={1,2,3,4,5,6,7},xticklabels={2DGS,Alicevision,PGSR,GGGS,NeRF2Mesh,SuGaR,3DGStoPC},xticklabel style={rotate=0},ylabel={Average Sentiment}]
				\addplot+[mark options={draw=black,fill=green},color=green]table[x=GPA,y=Leaf Color,col sep=comma]{plots/Userstudydata.csv};
				\addplot+[mark options={draw=black,fill=green},color=blue]table[x=GPA,y=Stem Quality,col sep=comma]{plots/Userstudydata.csv};
				\addplot+[mark options={draw=black,fill=green},color=yellow]table[x=GPA,y=Veins Visibility,col sep=comma]{plots/Userstudydata.csv};
				\addplot+[mark options={draw=black,fill=black},color=brown]table[x=GPA,y=Soil Detail,col sep=comma]{plots/Userstudydata.csv};
				\addplot+[mark options={draw=black,fill=brown},color=cyan]table[x=GPA,y=Branch Quality,col sep=comma]{plots/Userstudydata.csv};
				\legend{Leaf Color,Stem Quality,Veins Visibility,Soil Details,Branch Quality}
			\end{axis}
		\end{tikzpicture}
		\caption{The Average Sentiment scores for each of the considered aspects from the User-study can be observed for all 7 pipelines.}
		\label{fig:UserstudyBars}
	\end{figure*}
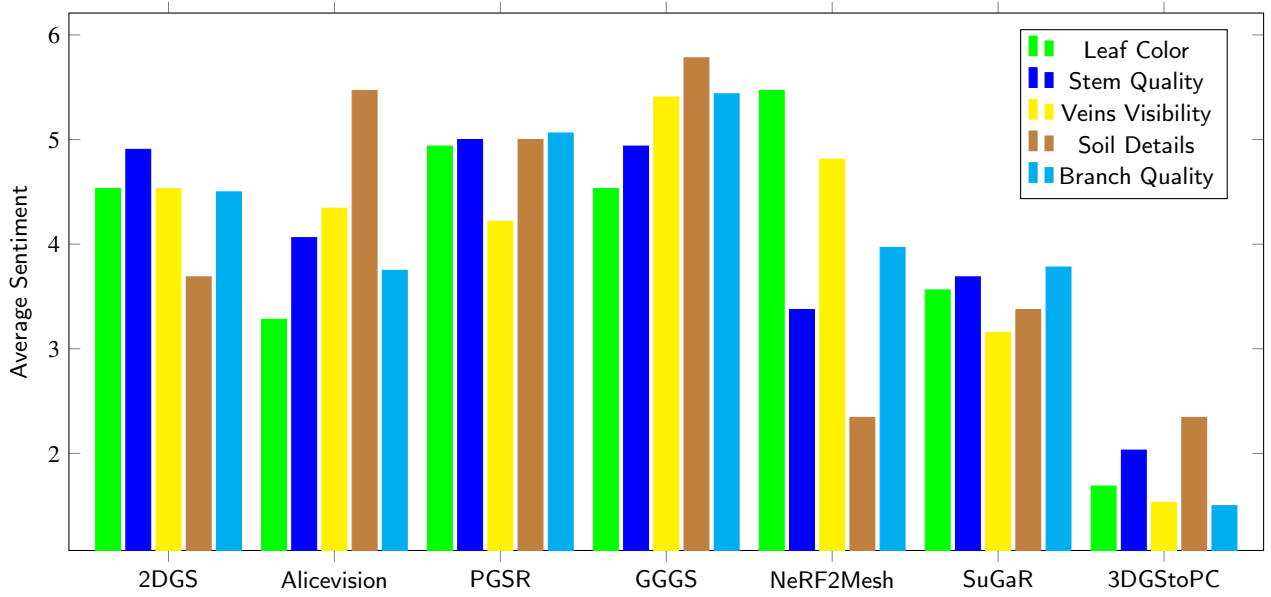

	\subsection{Result Analysis}
\begin{figure}
		\centering
		\begin{tikzpicture}
			\begin{axis}[ xtick={1,2,3,4,5,6,7},xticklabels={2DGS,Alicevision,PGSR,GGGS,NeRF2Mesh,SuGaR,3DGStoPC},,xticklabel style={rotate=45},ylabel={Final Computed Area}]
				\addplot+[only marks, mark options={draw=black,fill=red},color=red]table[x=GPA,y=AreaValue
				,col sep=comma]{plots/Spyder.csv};
			\end{axis}
		\end{tikzpicture}
		\caption{Area values computed on the radar chart, which are directly proportional to overall preference rankings (The higher the area, the higher the preference)}
		\label{fig:Areaval}
	\end{figure}
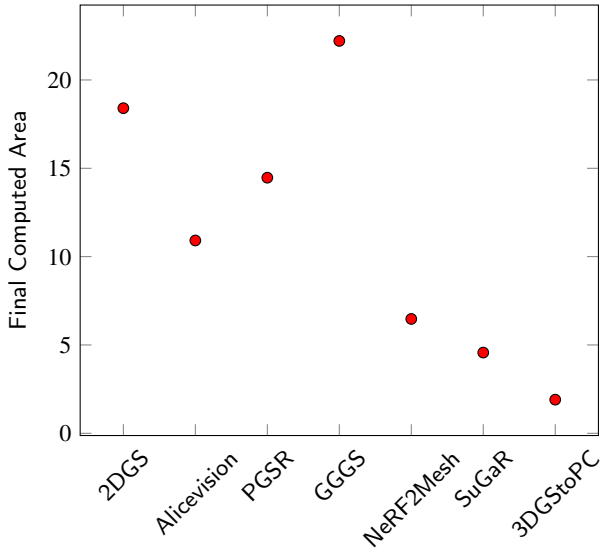

	Further summarizing the qualitative metrics, the Chamfer distance, PSNR, SSIM, and LPIPS were averaged across all 13 samples to obtain a single quantity for each pipeline.
	Based on this quantity, all the pipelines were arranged in 4 separate ranking orders for the corresponding metrics.
	Combining them with the User-study final ranking orders, we represented them via 5 dimensions on a radar chart, as in Figure \ref{fig:spiderweb}.
	The points on this radar chart are marked as per the rankings instead of the true metric values because the values being on different scales are not suitable for direct comparison.
	For clarity, it can be reiterated that the points further away from the center are higher in preference rankings (or better) as compared to the closer ones.
	We further computed the area covered by the spread of each pipeline on the radar chart as in Figure~\ref{fig:Areaval}.
	The pipeline with the highest area was given the maximum overall preference or the best rank.

    	\newcommand{\D}{5} 
	\newcommand{\U}{7} 
	
	\newdimen\R 
	\R=3.5cm 
	\newdimen\L 
	\L=4cm
	\newcommand{\A}{360/\D} 
	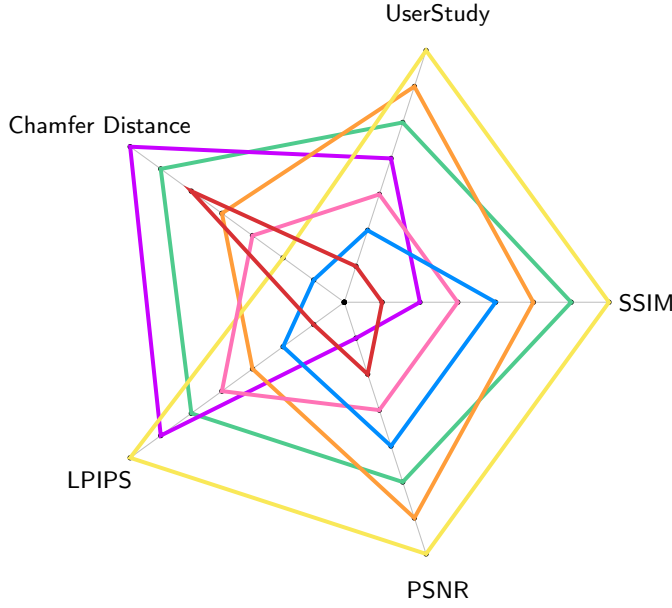
\begin{figure}
		\centering
		
		\begin{tikzpicture}[scale=1]
			\path (0:0cm) coordinate (O); 
			
			\foreach \X in {1,...,\D}{
				\draw [opacity=0.25] (\X*\A:0) -- (\X*\A:\R);
			}
			
			\foreach \Y in {0,...,\U}{
				\foreach \X in {1,...,\D}{
					\path (\X*\A:\Y*\R/\U) coordinate (D\X-\Y);
					\fill (D\X-\Y) circle (1pt);
				}
				\draw [opacity=0] (0:\Y*\R/\U) \foreach \X in {1,...,\D}{
					-- (\X*\A:\Y*\R/\U)
				} -- cycle;
			}
			
			
			\path (1*\A:\L) node (L1) {UserStudy};
			\path (2*\A:\L) node (L2) {Chamfer Distance};
			\path (3*\A:\L) node (L3) {LPIPS};
			\path (4*\A:\L) node (L4) {PSNR};
			\path (5*\A:\L) node (L5) {SSIM};
			
			\draw [color=CGreen,line width=1.5pt,opacity=1]
			(D1-5) --
			(D2-6) --
			(D3-5) --
			(D4-5) --
			(D5-6) -- cycle;
			
			\draw [color=CPurple,line width=1.5pt,opacity=1]
			(D1-4) --
			(D2-7) --
			(D3-6) --
			(D4-1) --
			(D5-2) -- cycle;
			
			\draw [color=COrange,line width=1.5pt,opacity=1]
			(D1-6) --
			(D2-4) --
			(D3-3) --
			(D4-6) --
			(D5-5) -- cycle;
			
			\draw [color=CYellow,line width=1.5pt,opacity=1]
			(D1-7) --
			(D2-2) --
			(D3-7) --
			(D4-7) --
			(D5-7) -- cycle;
			
			\draw [color=CPink,line width=1.5pt,opacity=1]
			(D1-3) --
			(D2-3) --
			(D3-4) --
			(D4-3) --
			(D5-3) -- cycle;
			
			\draw [color=MBlue,line width=1.5pt,opacity=1]
			(D1-2) --
			(D2-1) --
			(D3-2) --
			(D4-4) --
			(D5-4) -- cycle;
			
			\draw [color=CRed,line width=1.5pt,opacity=1]
			(D1-1) --
			(D2-5) --
			(D3-1) --
			(D4-2) --
			(D5-1) -- cycle;
			
		\end{tikzpicture}
		\caption{This radar chart summarizes the preference rankings of the methods based on five aspects: Userstudy, Chamfer Distance, LPIPS, SSIM, and PSNR. Points away from the center are better than the ones closer to the center on the preference scale. It can be seen that 2DGS is still one of the balanced choices. \textcolor{CRed}{\rule{7pt}{7pt}} 3DGStoPC \textcolor{MBlue}{\rule{7pt}{7pt}} SuGaR \textcolor{CPink}{\rule{7pt}{7pt}} NeRF2Mesh \textcolor{CGreen}{\rule{7pt}{7pt}} 2DGS \textcolor{CYellow}{\rule{7pt}{7pt}} GGGS \textcolor{CPurple}{\rule{7pt}{7pt}} Alicevision \textcolor{COrange}{\rule{7pt}{7pt}} PGSR}
		\label{fig:spiderweb}
	\end{figure}    
	
	\subsubsection{Observations}
	The GGGS pipeline is at the top of the ranking order, followed by 2DGS, PGSR, Alicevision, NeRF2Mesh, SuGaR, and 3DGS-to-PC in the same order.
	Between 2DGS and PGSR, the Chamfer distance, LPIPS, and SSIM were better for 2DGS, while PGSR had better PSNR and User ratings.
	Alicevision, being the representative of the traditional photogrammetry pipelines, ensured a low LPIPS value, indicating that the reconstruction was perceptually closer to humans, second to GGGS.
	NeRF2Mesh was the highly preferred method for accurate leaf colors as per the User Study.
	The photorealistic details of the veins on the leaves were very clear.
	SuGaR pipeline was one of the pioneers to perform reconstruction by aligning the Gaussians along a 2D surface, which was mastered by 2DGS later.
	But SuGaR additionally generates the UV textures, which is not the case with the latest Gaussian-based methods, as they just work with the vertex colors only.
	
\begin{figure}
		\centering
		\begin{subfigure}{0.23\textwidth}
			\centering
			\includegraphics[width=\textwidth]{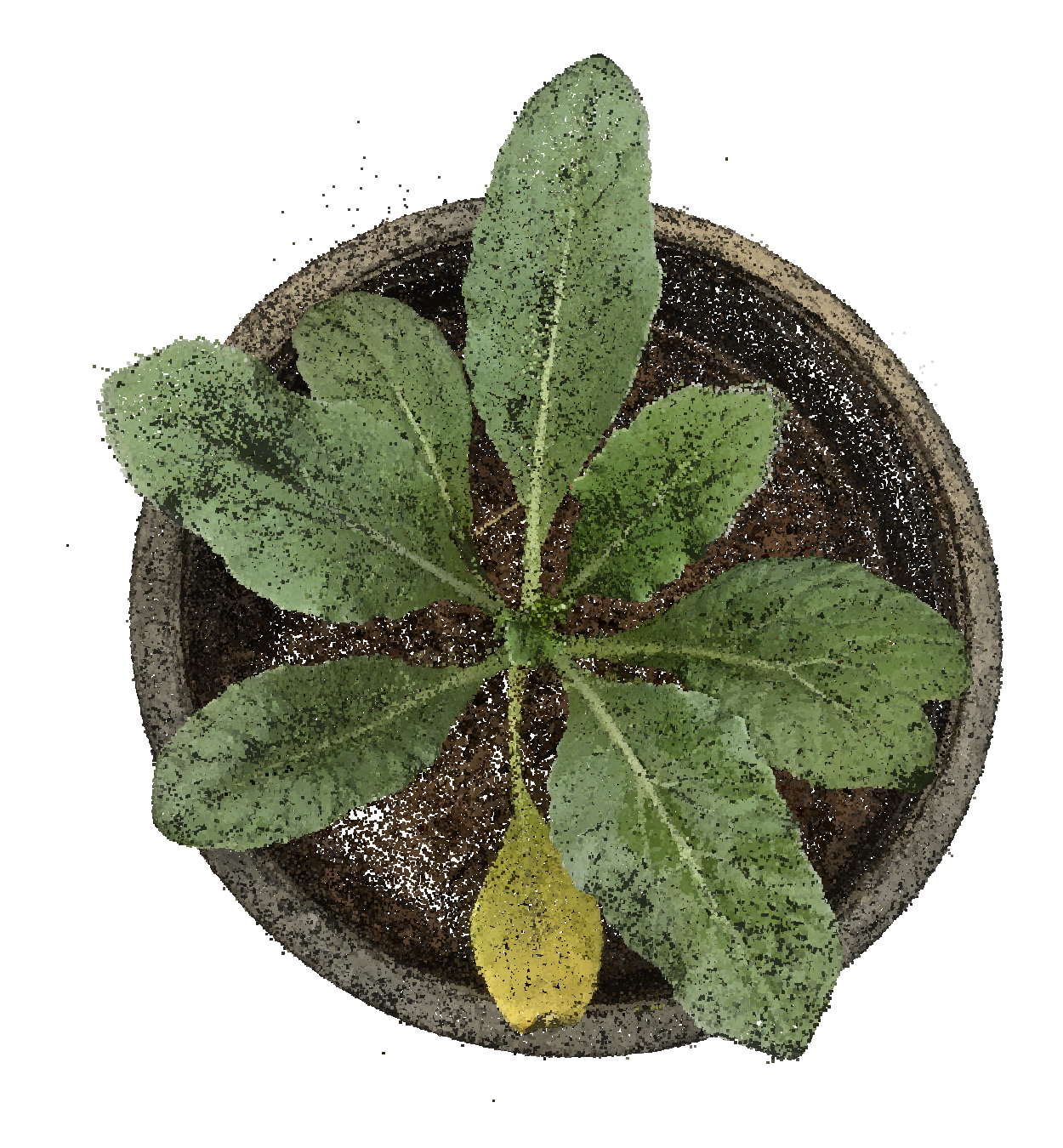}
			\caption{Cleaned Point Cloud}
		\end{subfigure}
		\hfill
		\begin{subfigure}{0.23\textwidth}
			\centering
			\includegraphics[width=\textwidth]{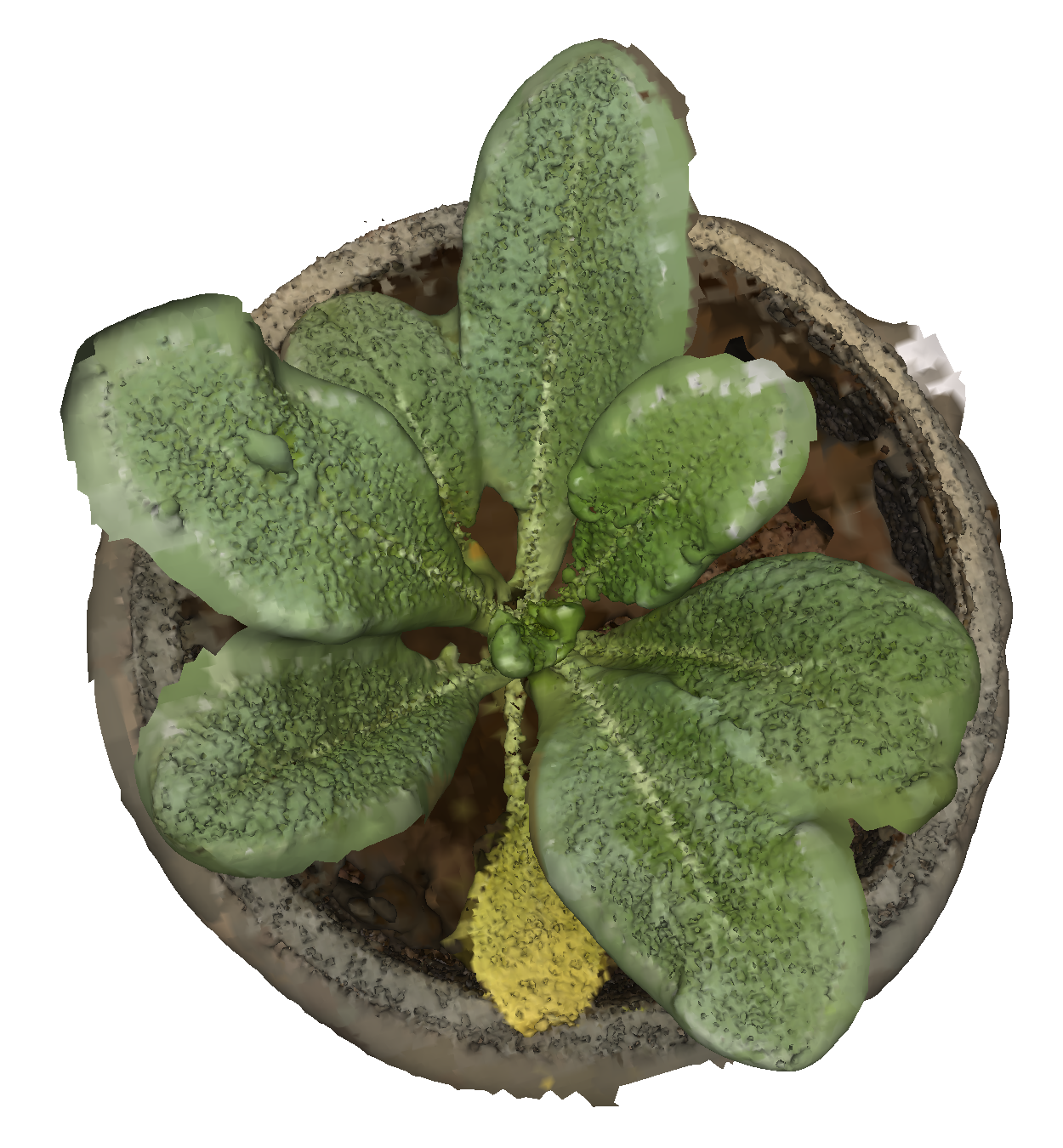}
			\caption{Cleaned Mesh}
		\end{subfigure}
		\caption{3DGS-to-PC Pipeline Example - Day 9}
		\label{fig:pcandmesh}
	\end{figure}

	The mesh reconstruction of the 3DGS-to-PC method was poor, but its point cloud reconstruction abilities were good, as can be observed via the Chamfer distance values, which was better than even the GGGS, PGSR, and NeRF2Mesh pipelines.
	However, the Chamfer distance metric considers the point clouds only, rather than the entire mesh.
	Consequently, the final mesh in this case was poorly reconstructed.
	This can be seen in one of the samples (Day-9 of our dataset) as in Figure \ref{fig:pcandmesh}.
	Hence, this pipeline can be a good candidate for the quick reconstruction of a point cloud from the 3DGS output if the point cloud is the only requirement.
    	
	\subsubsection{Visual Results}
\begin{figure}
		\centering
		\begin{subfigure}[h]{0.23\textwidth}
			\centering
			\includegraphics[width=1\textwidth]{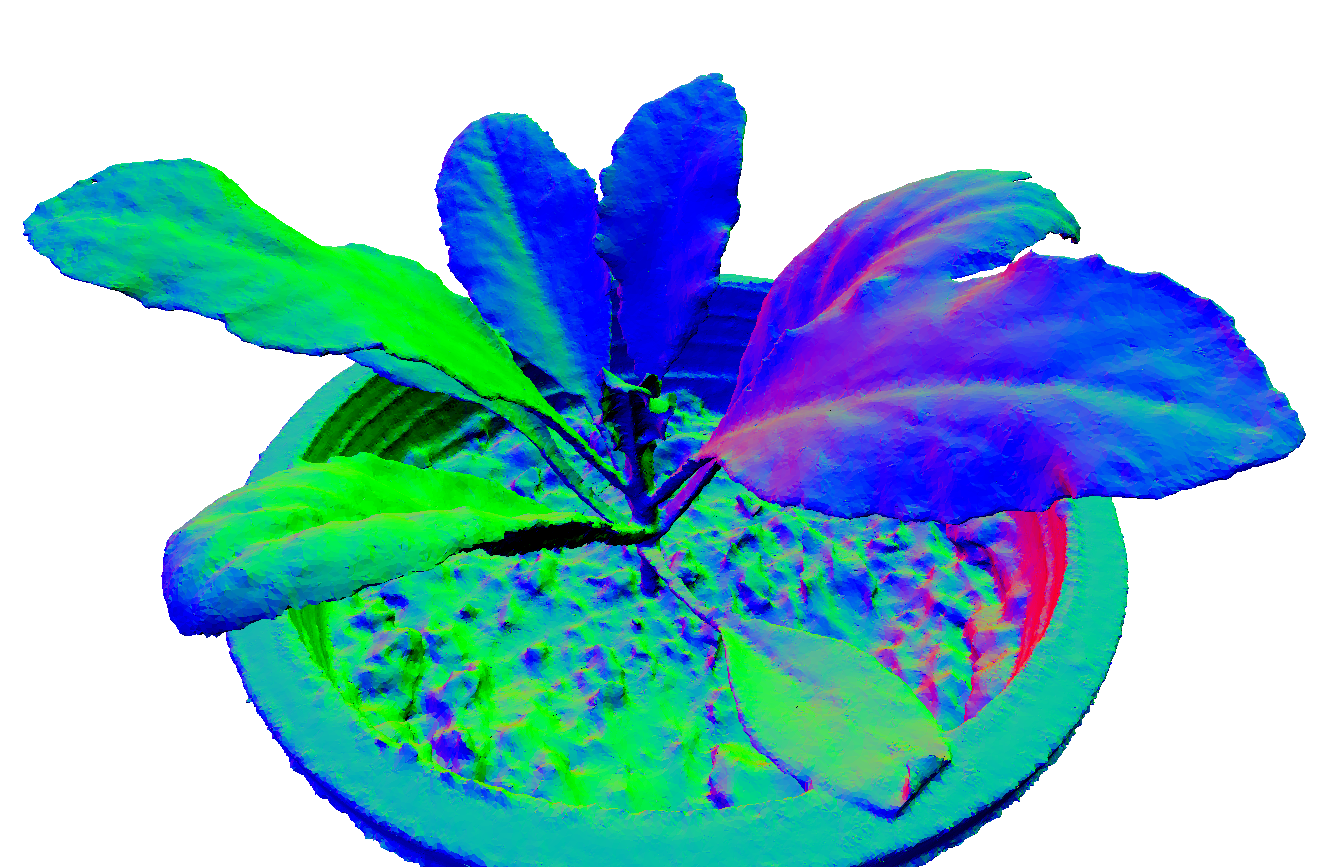}
			\caption{Alicevision Meshroom}
		\end{subfigure}
		\hfill
		\begin{subfigure}[h]{0.23\textwidth}
			\centering
			\includegraphics[width=1\textwidth]{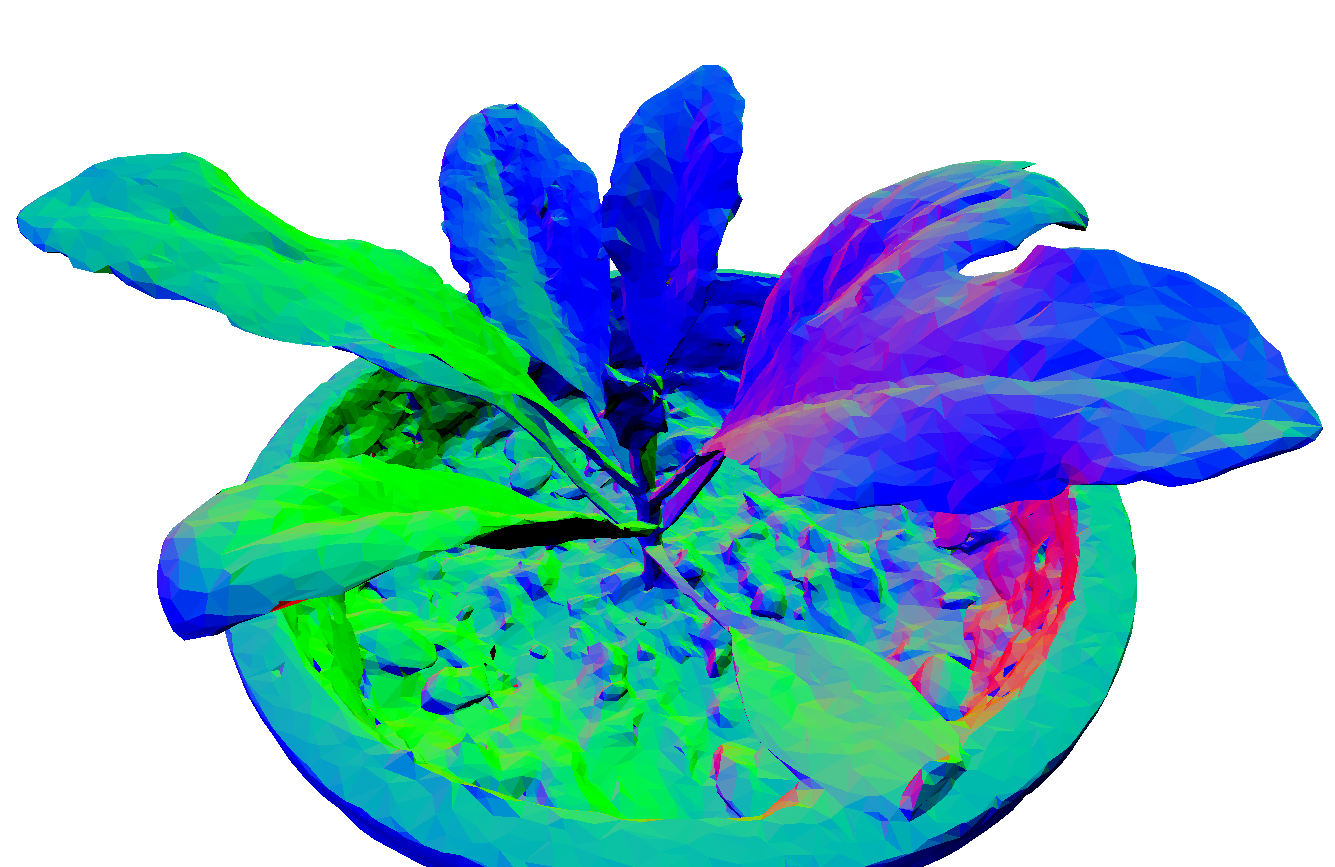}
			\caption{SuGaR}
		\end{subfigure}
		\begin{subfigure}[h]{0.23\textwidth}
			\centering
			\includegraphics[width=1\textwidth]{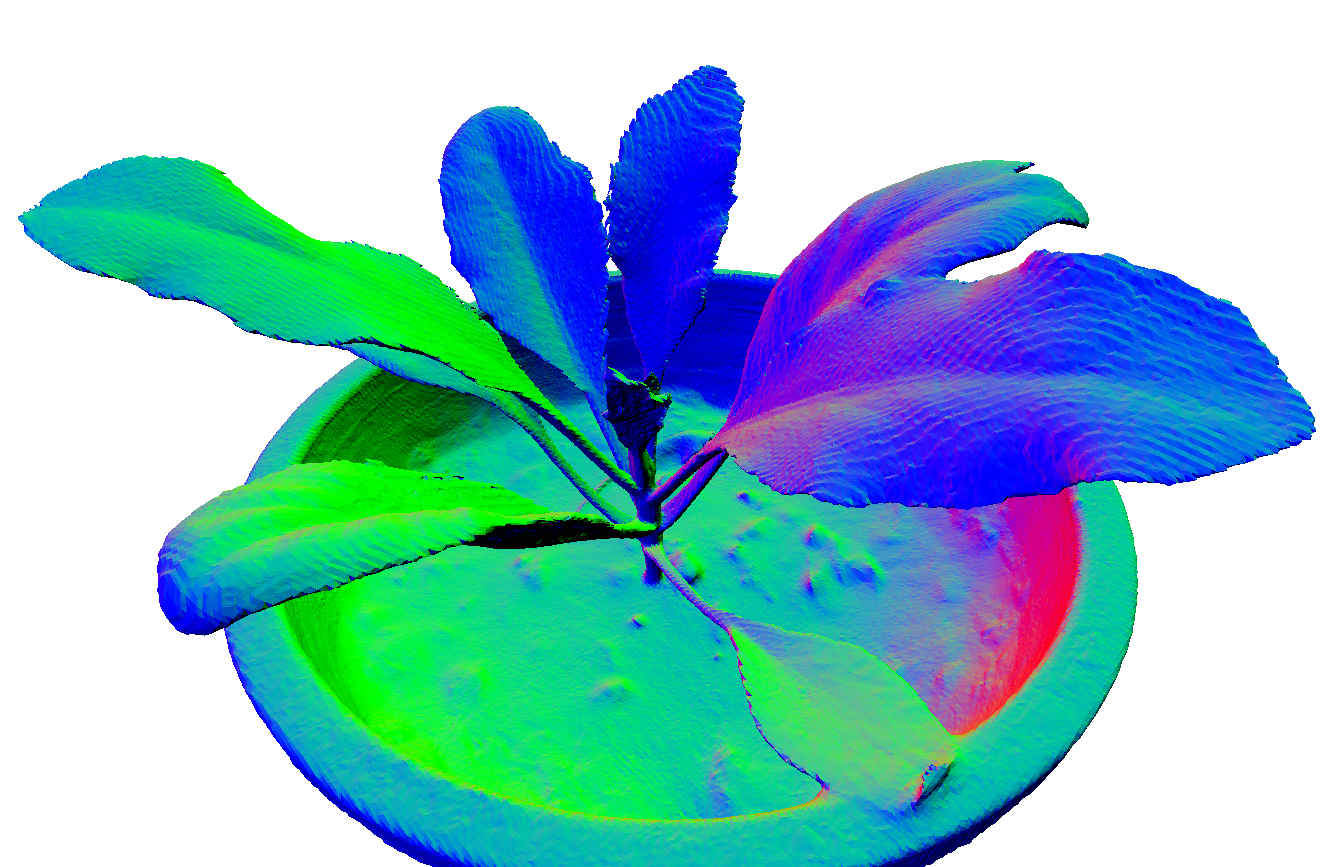}
			\caption{2D Gaussian Splats}
		\end{subfigure}
		\hfill
		\begin{subfigure}[h]{0.23\textwidth}
			\centering
			\includegraphics[width=1\textwidth]{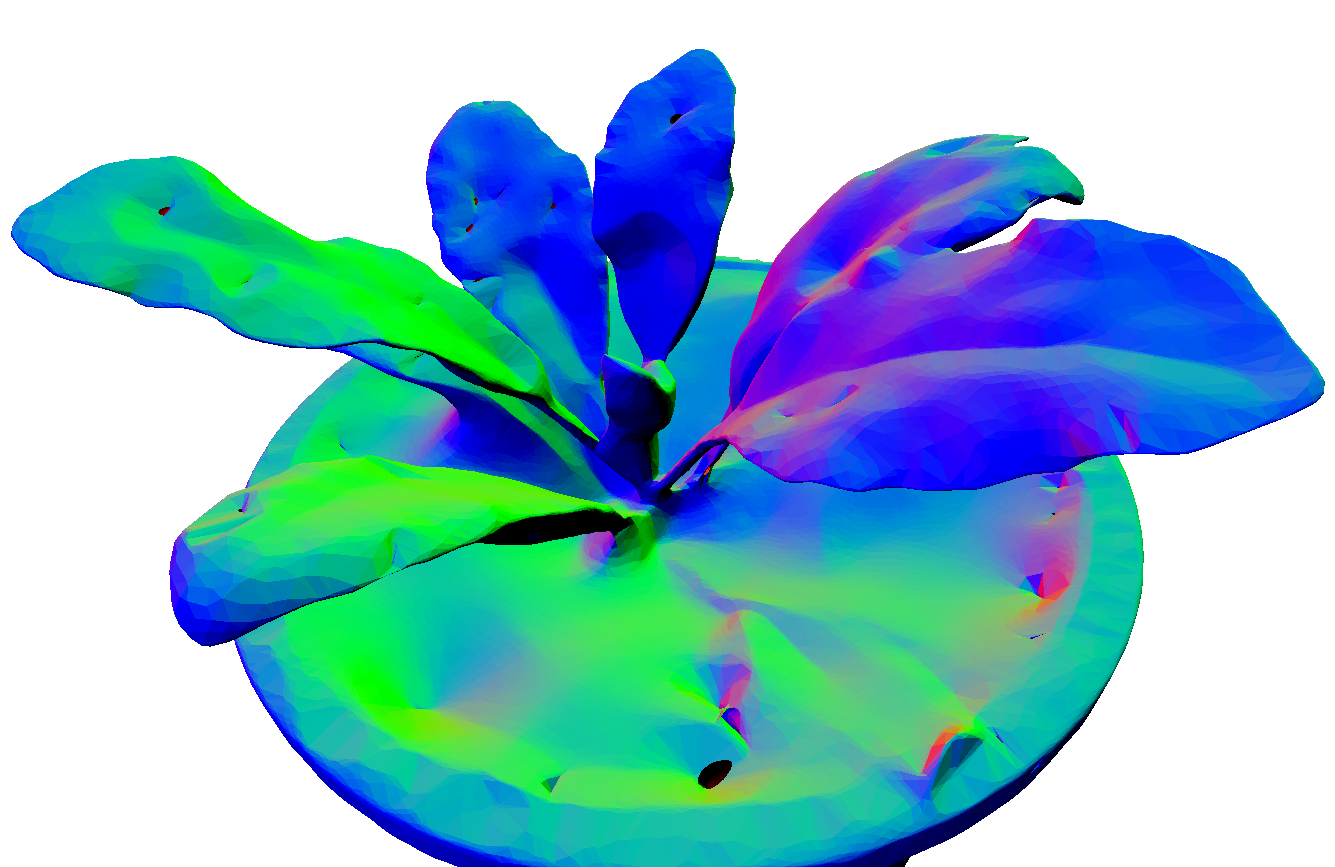}
			\caption{NeRF2Mesh}
		\end{subfigure}
		\begin{subfigure}[h]{0.23\textwidth}
			\centering
			\includegraphics[width=1\textwidth]{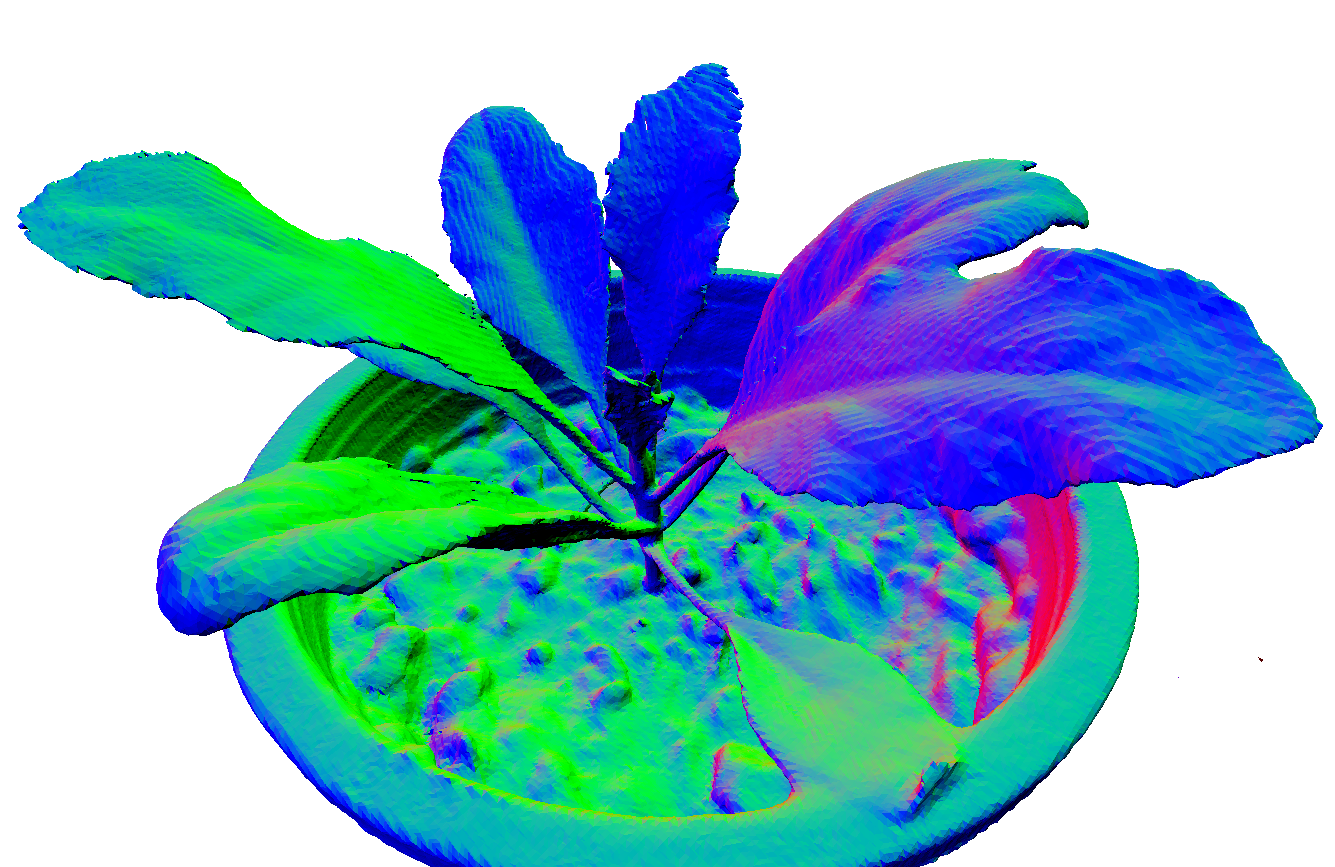}
			\caption{PGSR}
		\end{subfigure}
		\hfill
		\begin{subfigure}[h]{0.23\textwidth}
			\centering
			\includegraphics[width=1\textwidth]{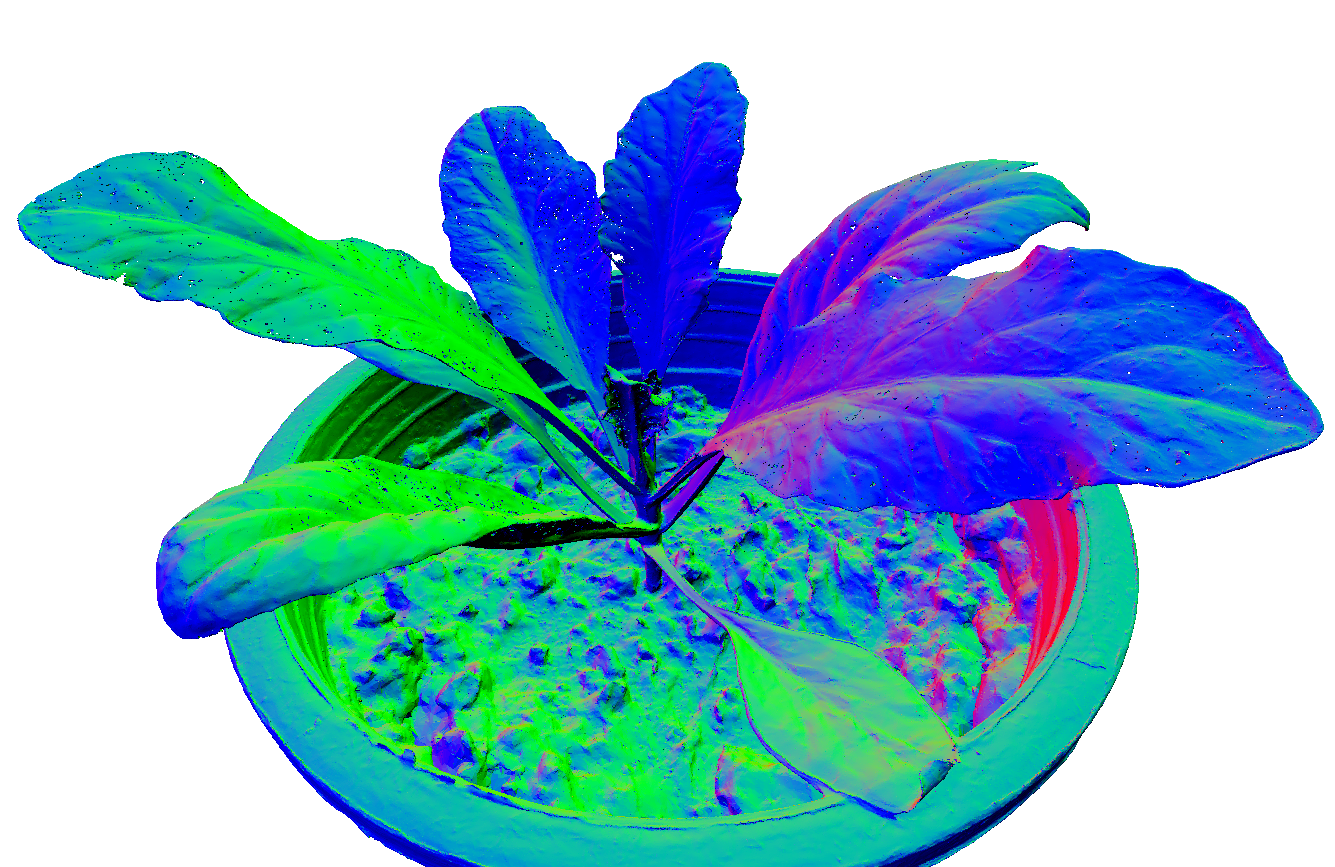}
			\caption{GGGS}
		\end{subfigure}
		\caption{Normal Maps from the Front View of Day 9 Cauliflower for all the Pipelines.}
		\label{fig:Front View Normals}
	\end{figure}

\begin{figure}
		\centering
		\begin{subfigure}[h]{0.23\textwidth}
			\centering
			\includegraphics[width=1\textwidth]{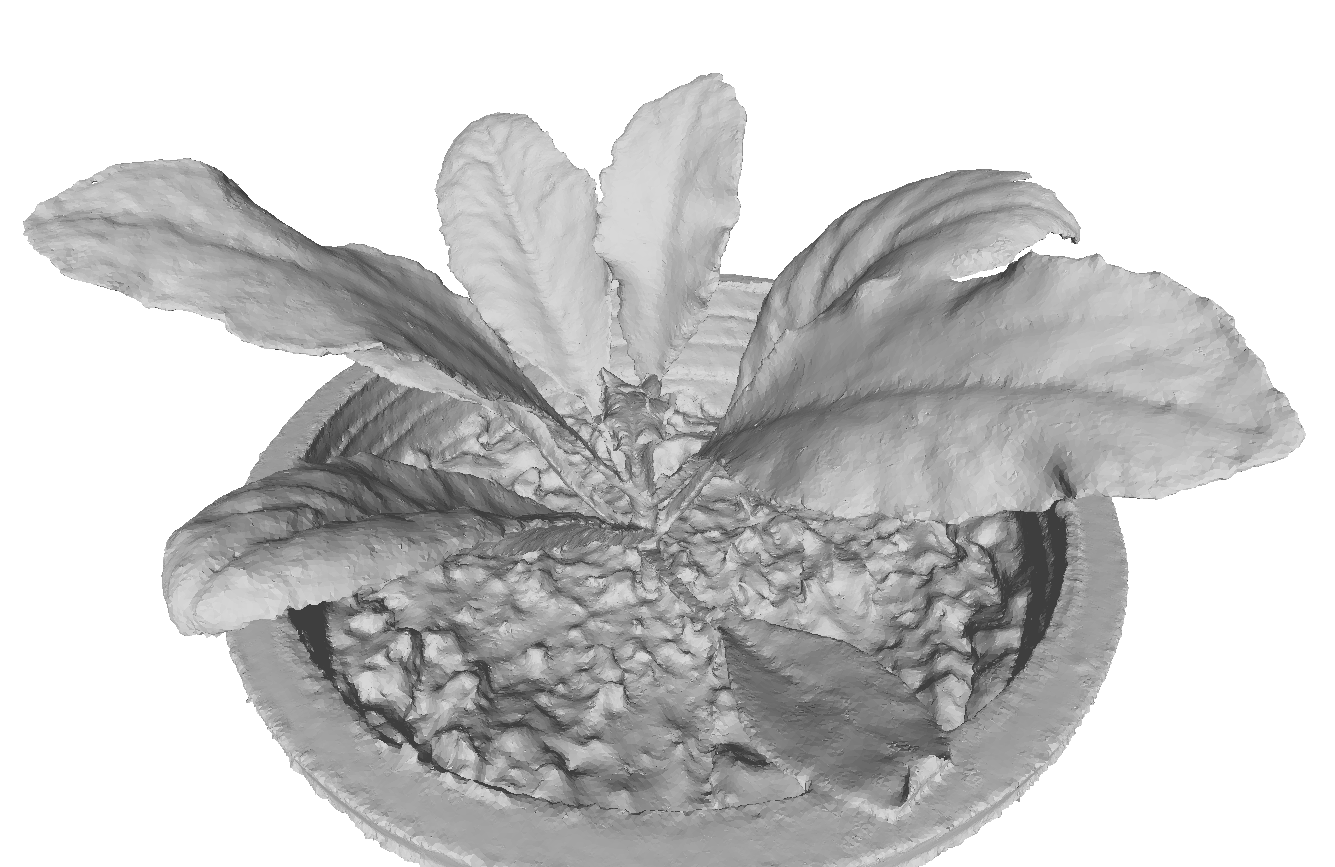}
			\caption{Alicevision Meshroom}
		\end{subfigure}
		\hfill
		\begin{subfigure}[h]{0.23\textwidth}
			\centering
			\includegraphics[width=1\textwidth]{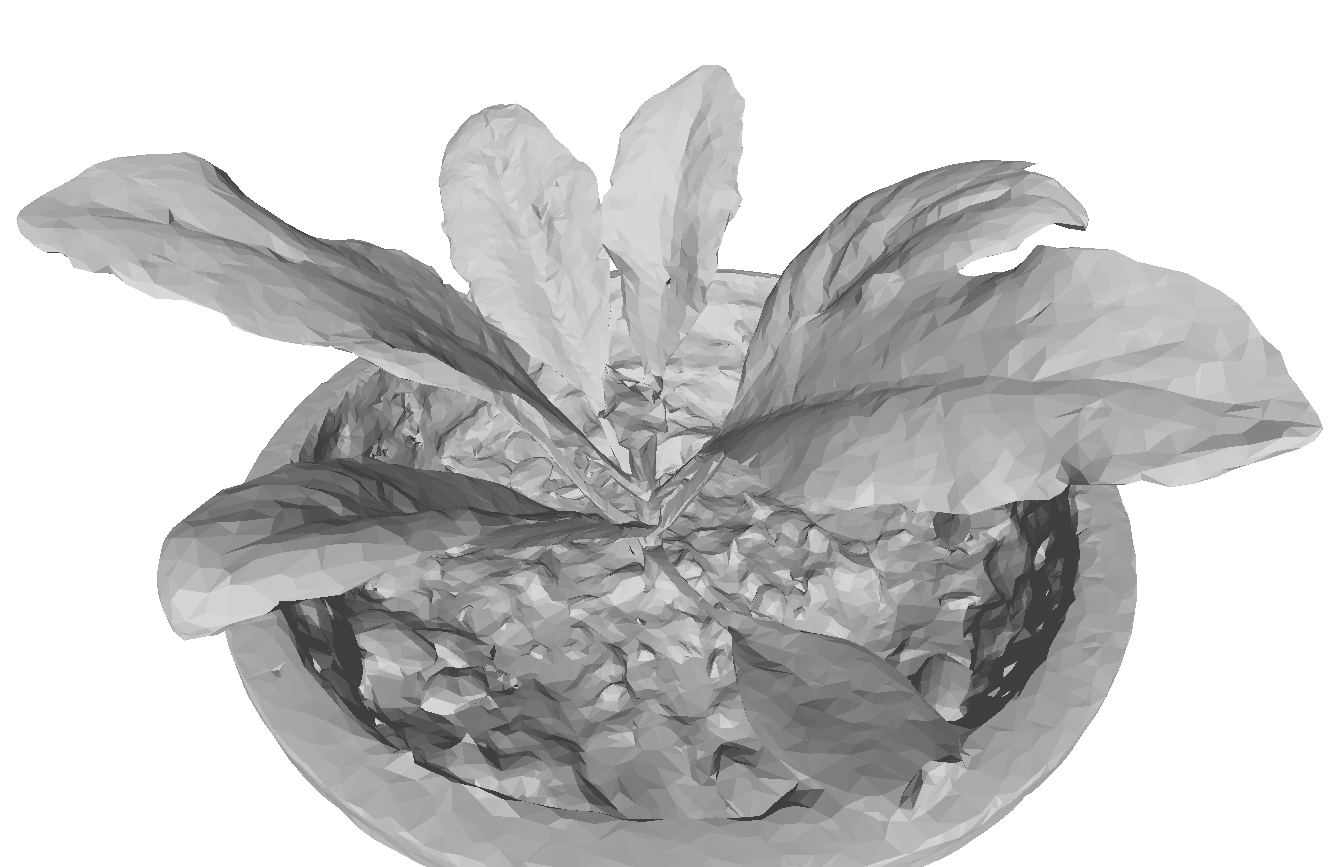}
			\caption{SuGaR}
		\end{subfigure}
		\begin{subfigure}[h]{0.23\textwidth}
			\centering
			\includegraphics[width=1\textwidth]{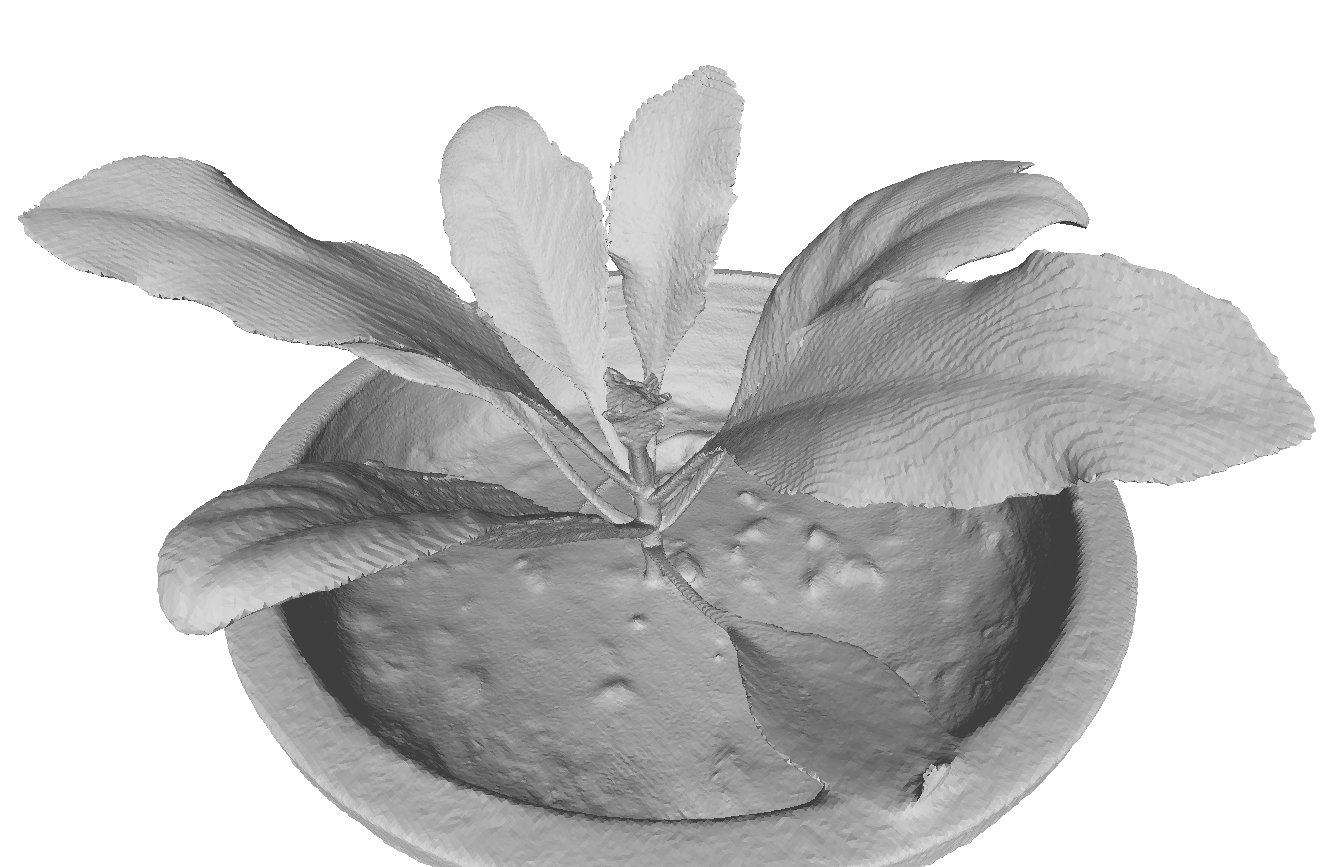}
			\caption{2D Gaussian Splats}
		\end{subfigure}
		\hfill
		\begin{subfigure}[h]{0.23\textwidth}
			\centering
			\includegraphics[width=1\textwidth]{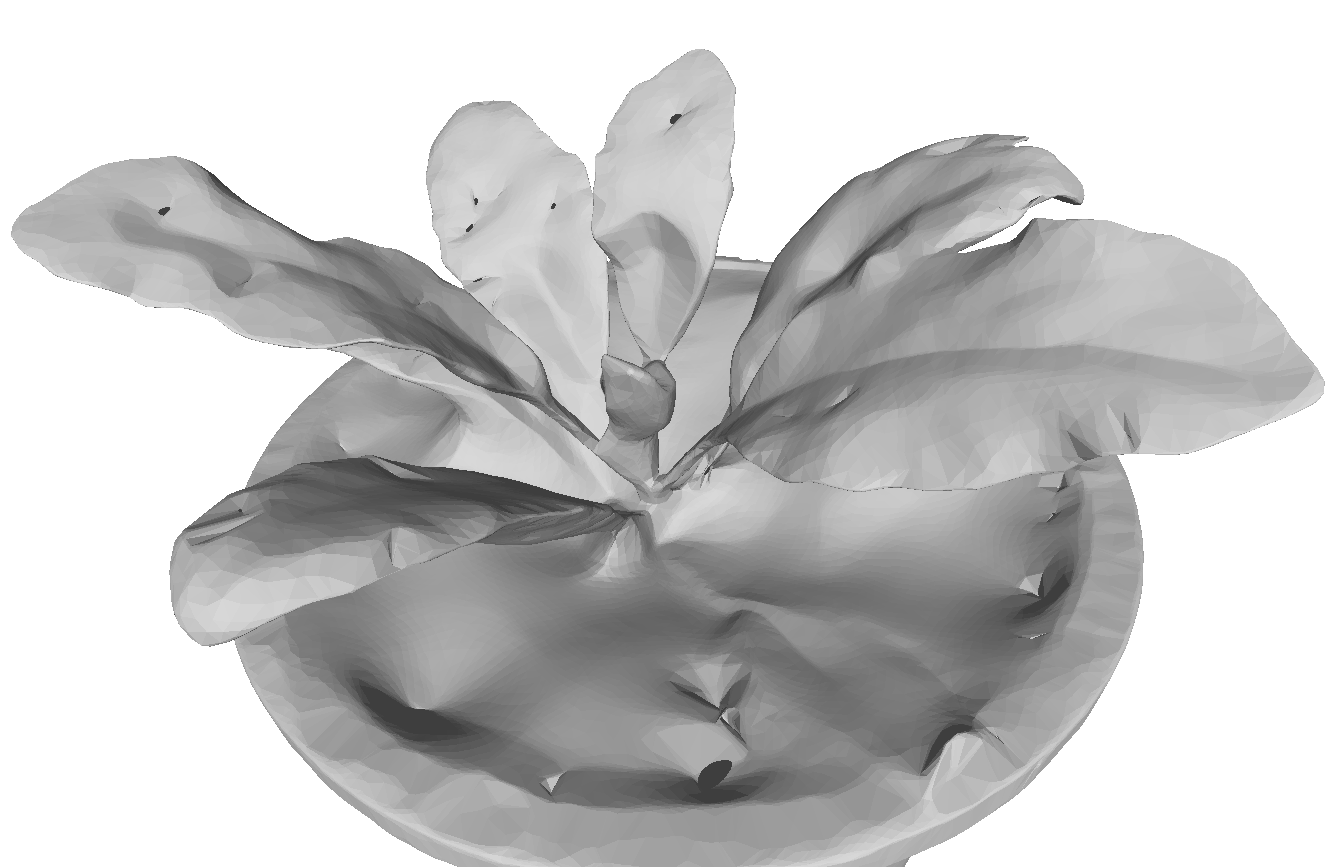}
			\caption{NeRF2Mesh}
		\end{subfigure}
		\begin{subfigure}[h]{0.23\textwidth}
			\centering
			\includegraphics[width=1\textwidth]{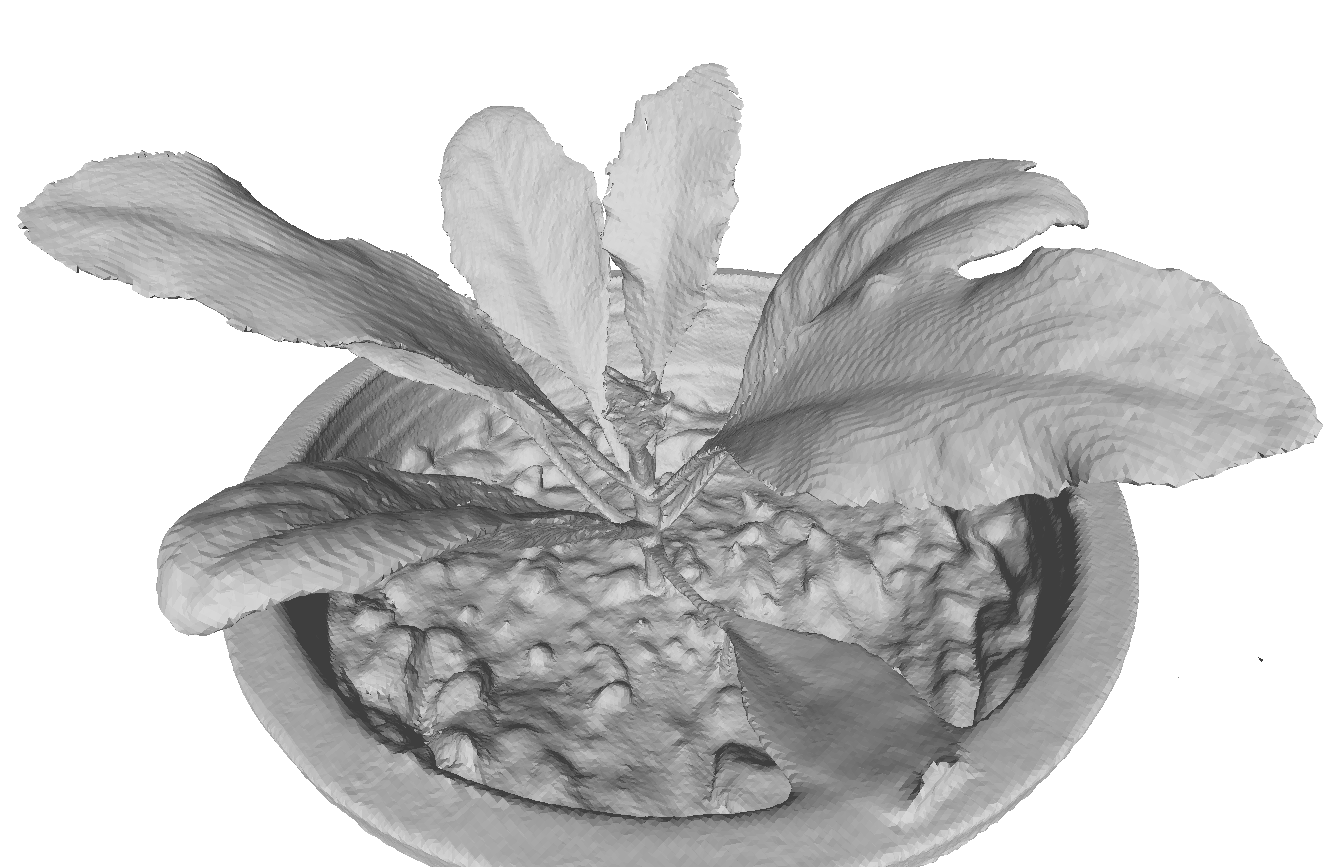}
			\caption{PGSR}
		\end{subfigure}
		\hfill
		\begin{subfigure}[h]{0.23\textwidth}
			\centering
			\includegraphics[width=1\textwidth]{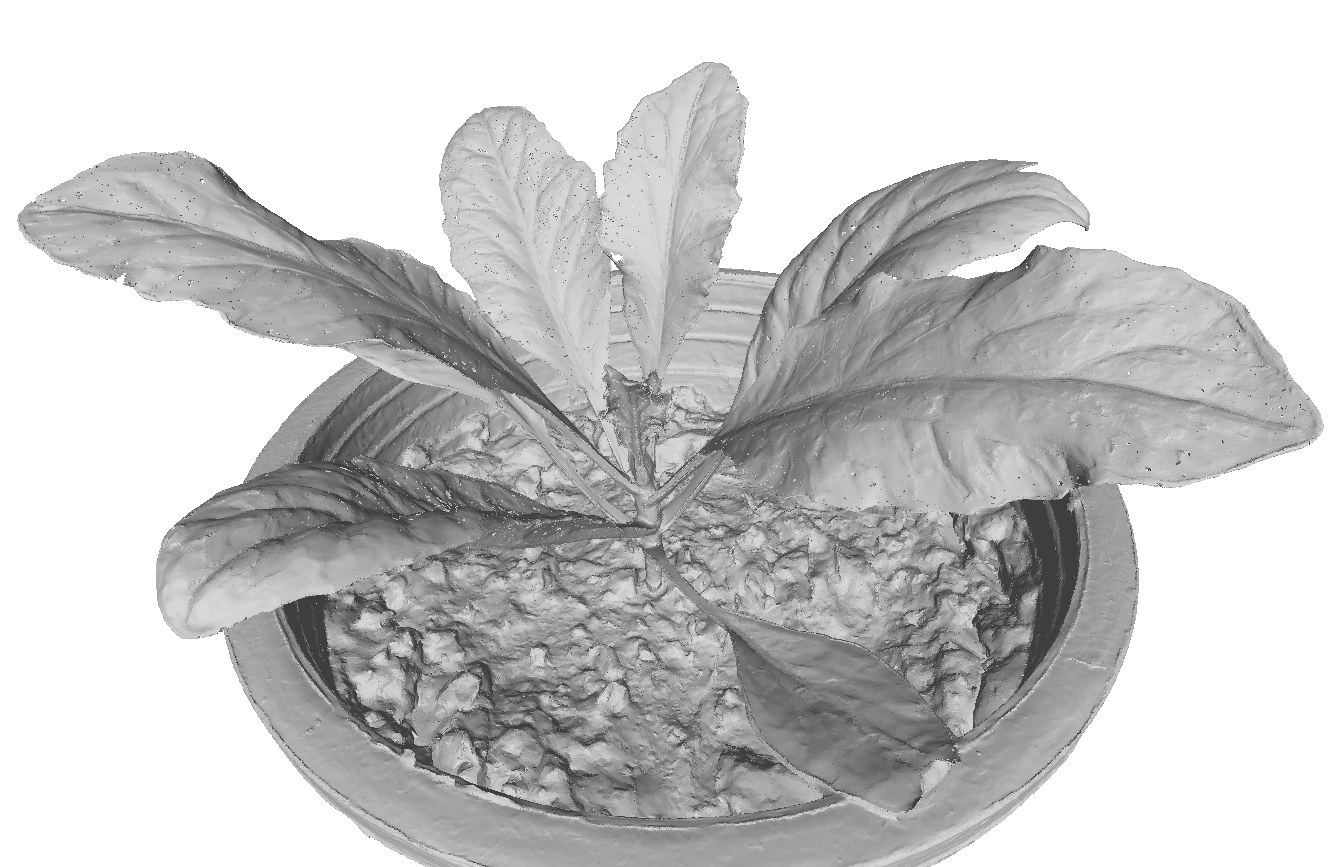}
			\caption{GGGS}
		\end{subfigure}
		\caption{Geometrical Front View of Day 9 Cauliflower for all the Pipelines}
		\label{fig:Grayscale Front View}
	\end{figure}

\begin{figure}
		\centering
		\begin{subfigure}[h]{0.23\textwidth}
			\centering
			\includegraphics[width=1\textwidth]{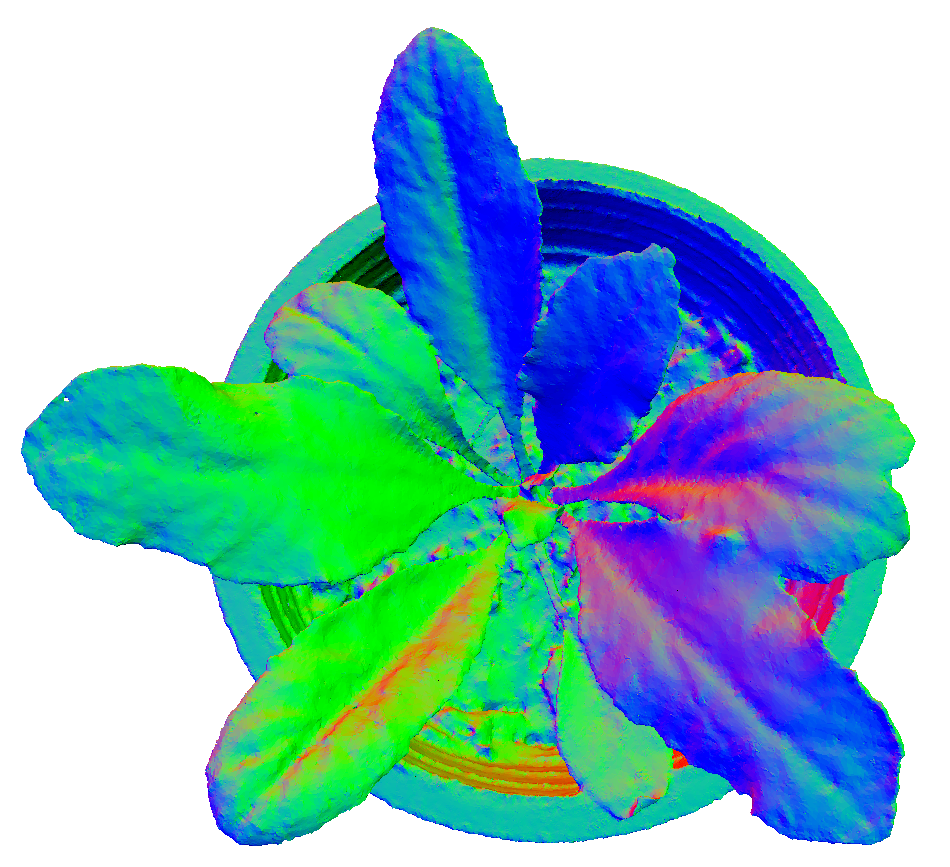}
			\caption{Alicevision Meshroom}
		\end{subfigure}
		\hfill
		\begin{subfigure}[h]{0.23\textwidth}
			\centering
			\includegraphics[width=1\textwidth]{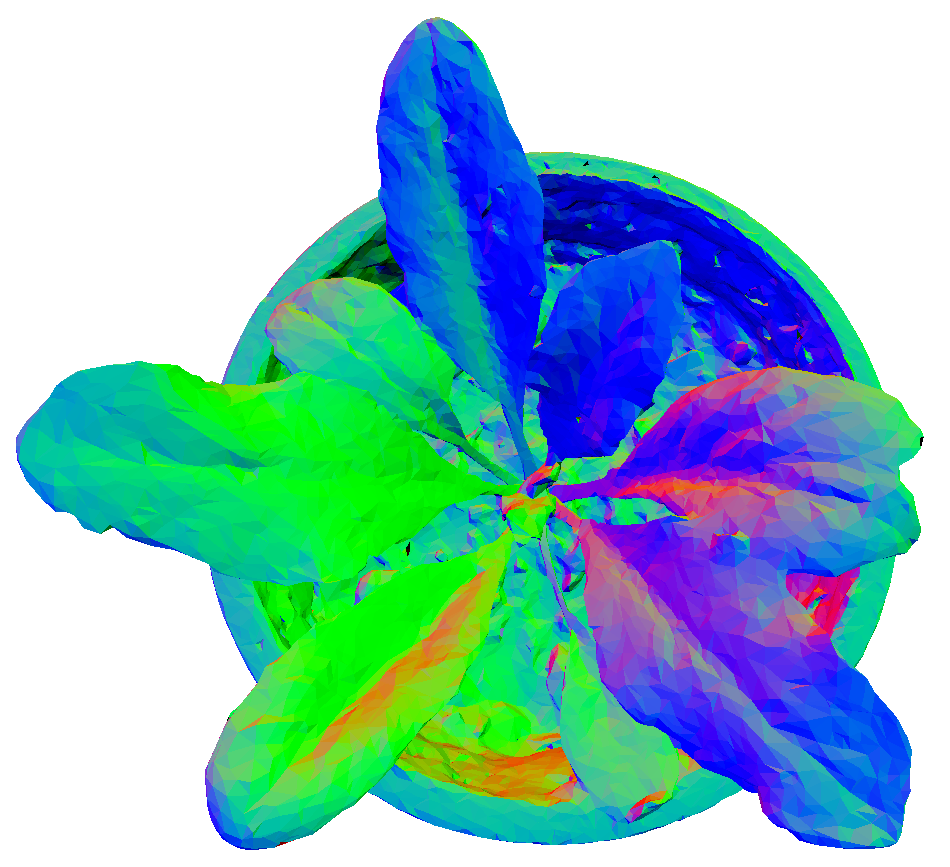}
			\caption{SuGaR}
		\end{subfigure}
		\begin{subfigure}[h]{0.23\textwidth}
			\centering
			\includegraphics[width=1\textwidth]{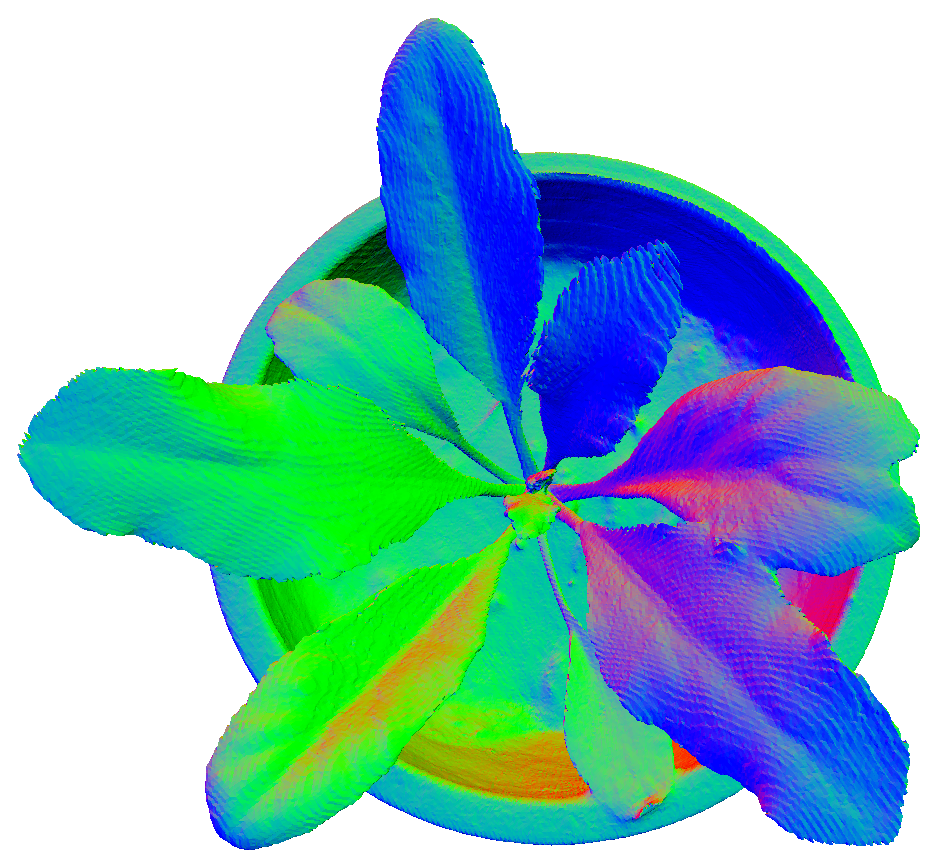}
			\caption{2D Gaussian Splats}
		\end{subfigure}
		\hfill
		\begin{subfigure}[h]{0.23\textwidth}
			\centering
			\includegraphics[width=1\textwidth]{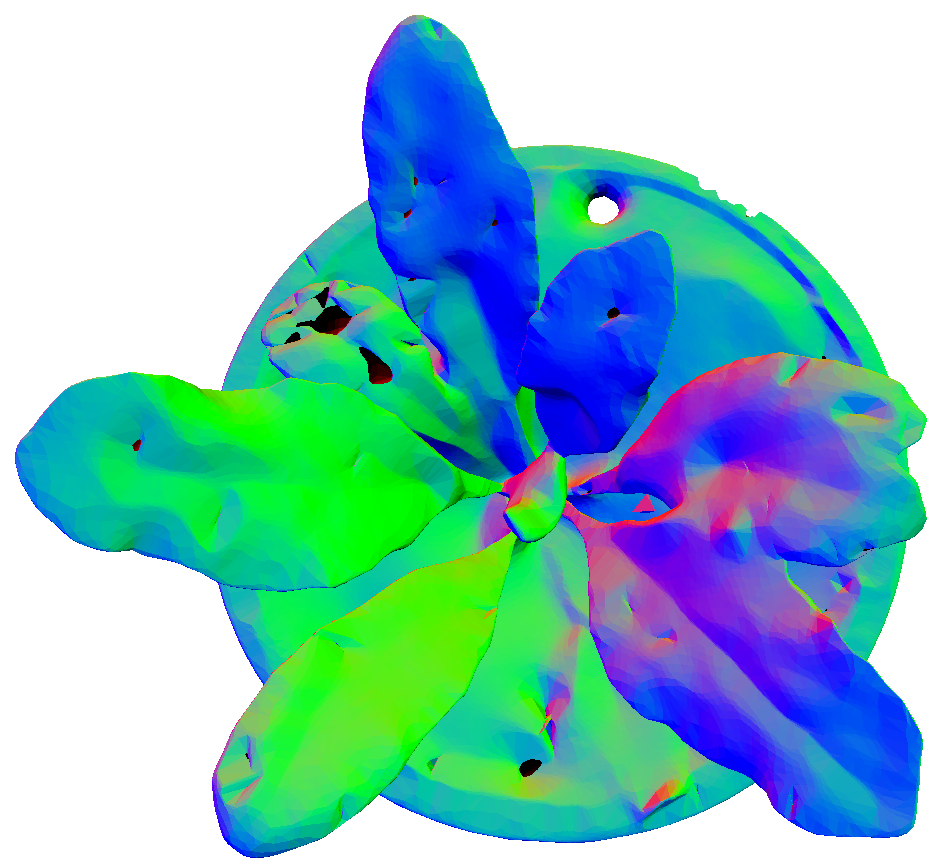}
			\caption{NeRF2Mesh}
		\end{subfigure}
		\begin{subfigure}[h]{0.23\textwidth}
			\centering
			\includegraphics[width=1\textwidth]{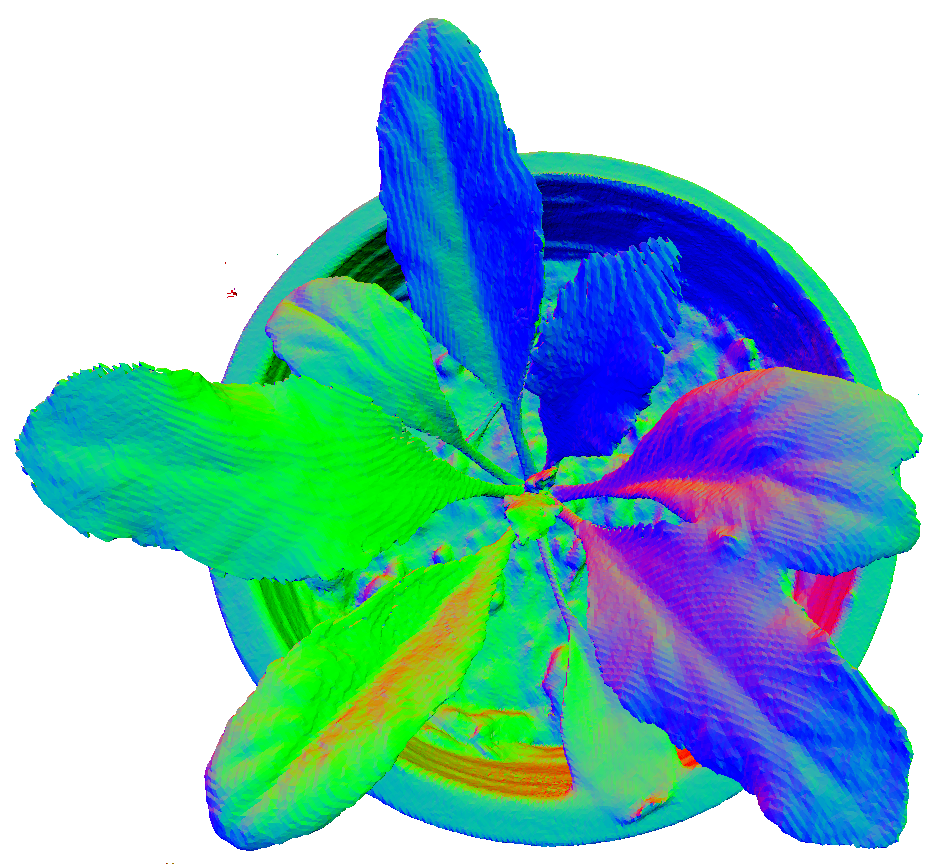}
			\caption{PGSR}
		\end{subfigure}
		\hfill
		\begin{subfigure}[h]{0.23\textwidth}
			\centering
			\includegraphics[width=1\textwidth]{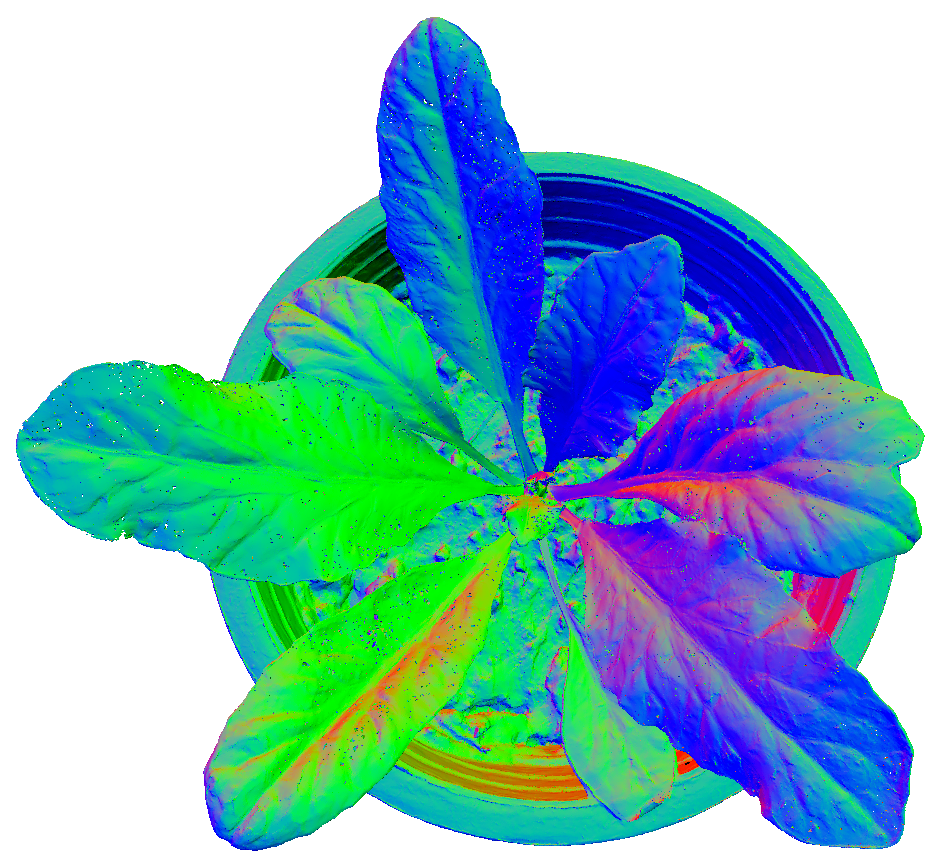}
			\caption{GGGS}
		\end{subfigure}
		\caption{Normal Maps from the Top View of Day 9 Cauliflower for all the Pipelines}
		\label{fig:Top View Normals}
	\end{figure}

\begin{figure}
		\centering
		\begin{subfigure}[h]{0.23\textwidth}
			\centering
			\includegraphics[width=1\textwidth]{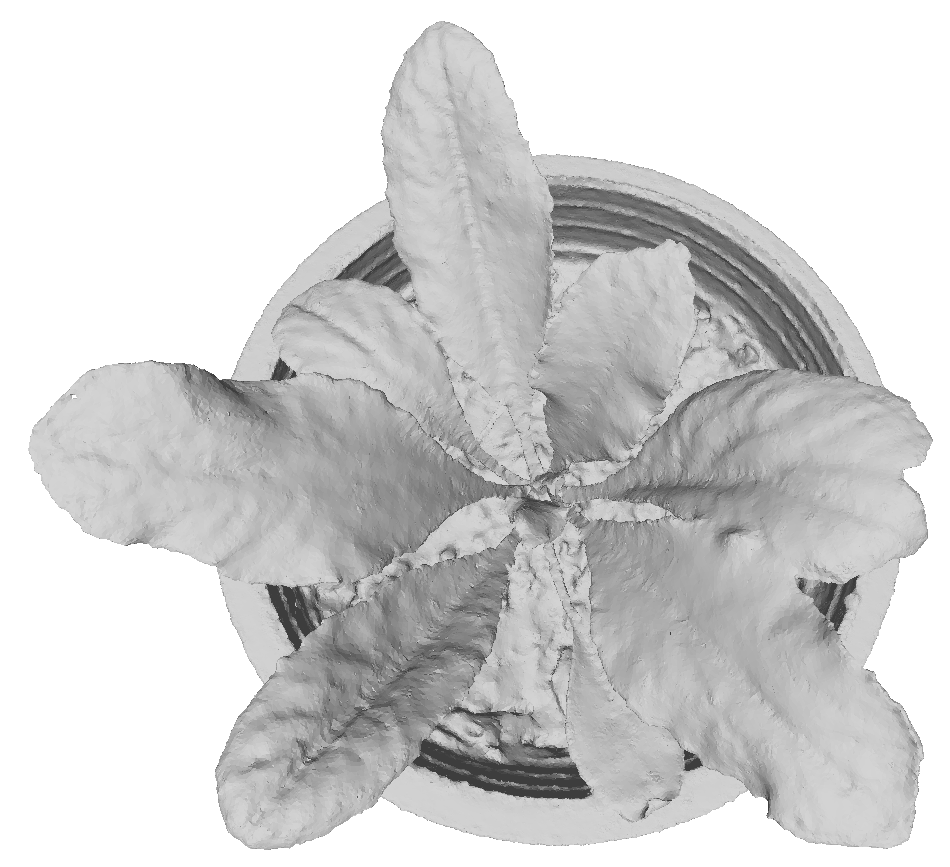}
			\caption{Alicevision Meshroom}
		\end{subfigure}
		\hfill
		\begin{subfigure}[h]{0.23\textwidth}
			\centering
			\includegraphics[width=1\textwidth]{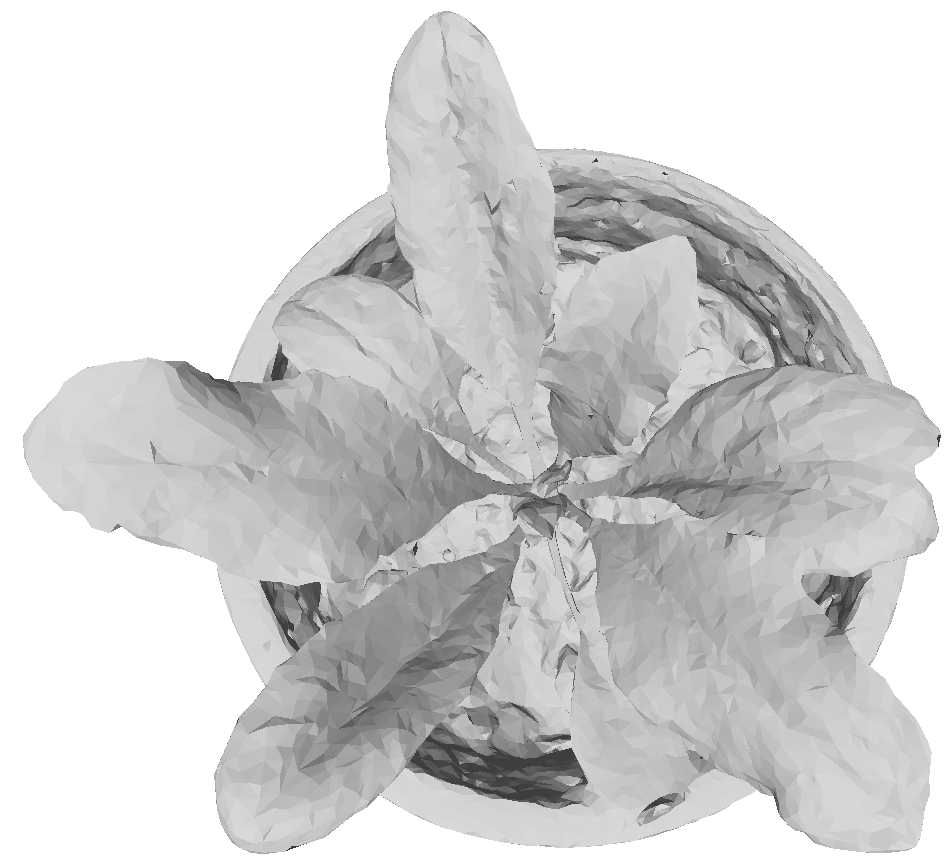}
			\caption{SuGaR}
		\end{subfigure}
		\begin{subfigure}[h]{0.23\textwidth}
			\centering
			\includegraphics[width=1\textwidth]{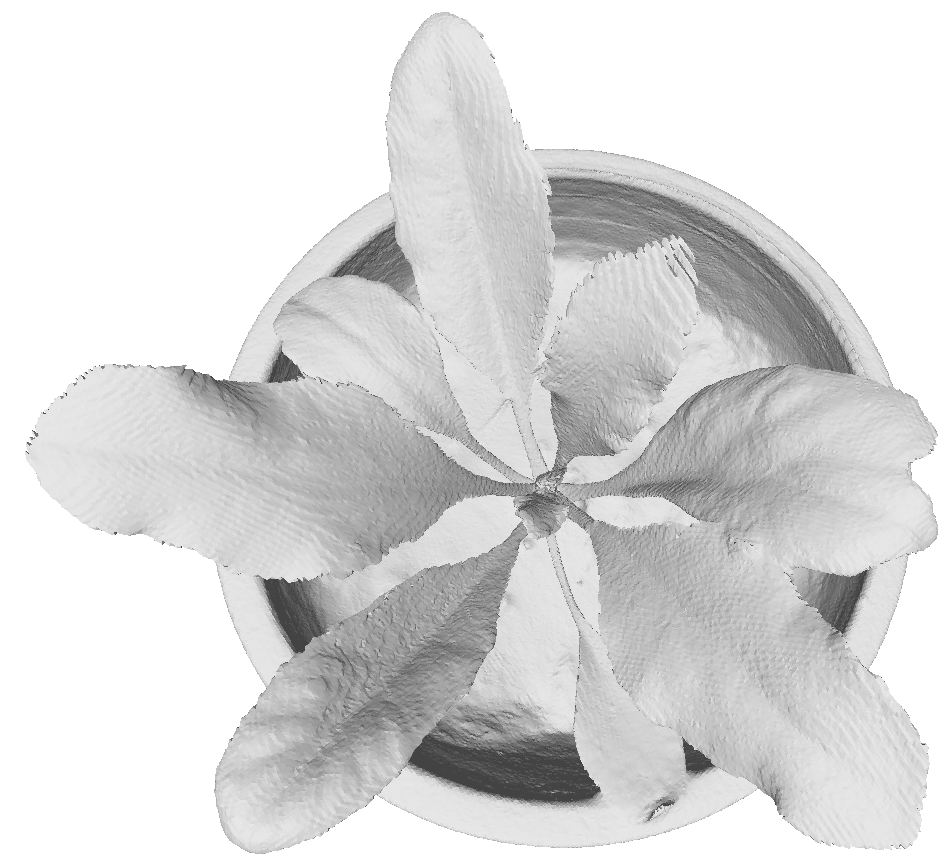}
			\caption{2D Gaussian Splats}
		\end{subfigure}
		\hfill
		\begin{subfigure}[h]{0.23\textwidth}
			\centering
			\includegraphics[width=1\textwidth]{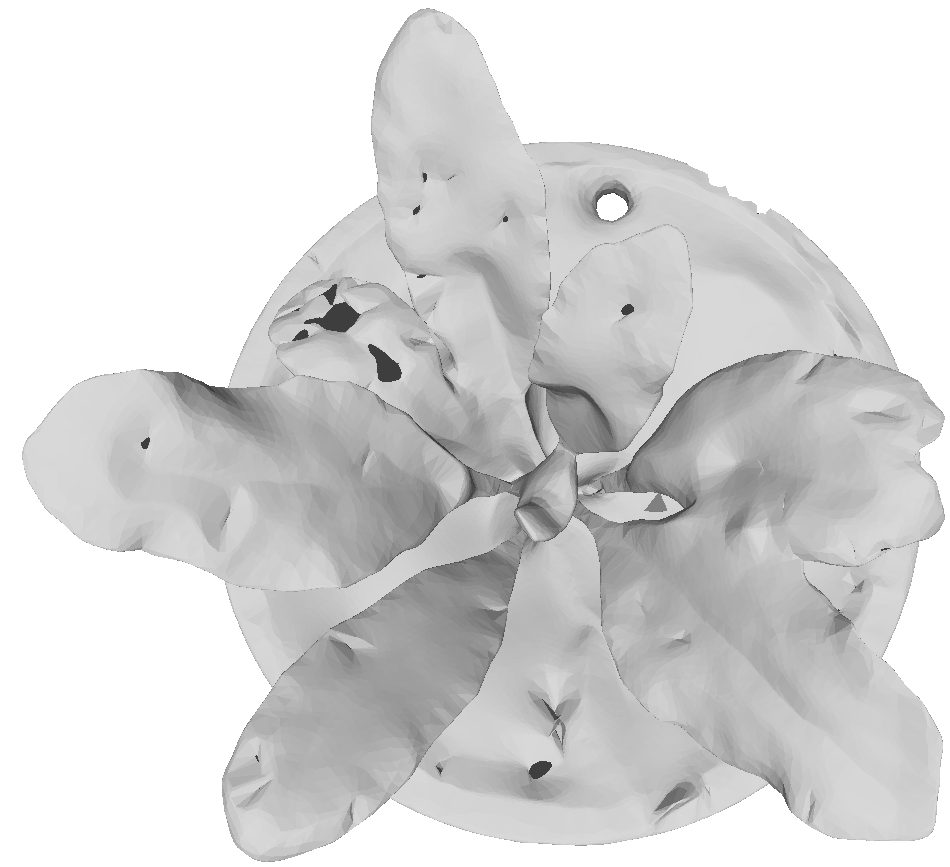}
			\caption{NeRF2Mesh}
		\end{subfigure}
		\begin{subfigure}[h]{0.23\textwidth}
			\centering
			\includegraphics[width=1\textwidth]{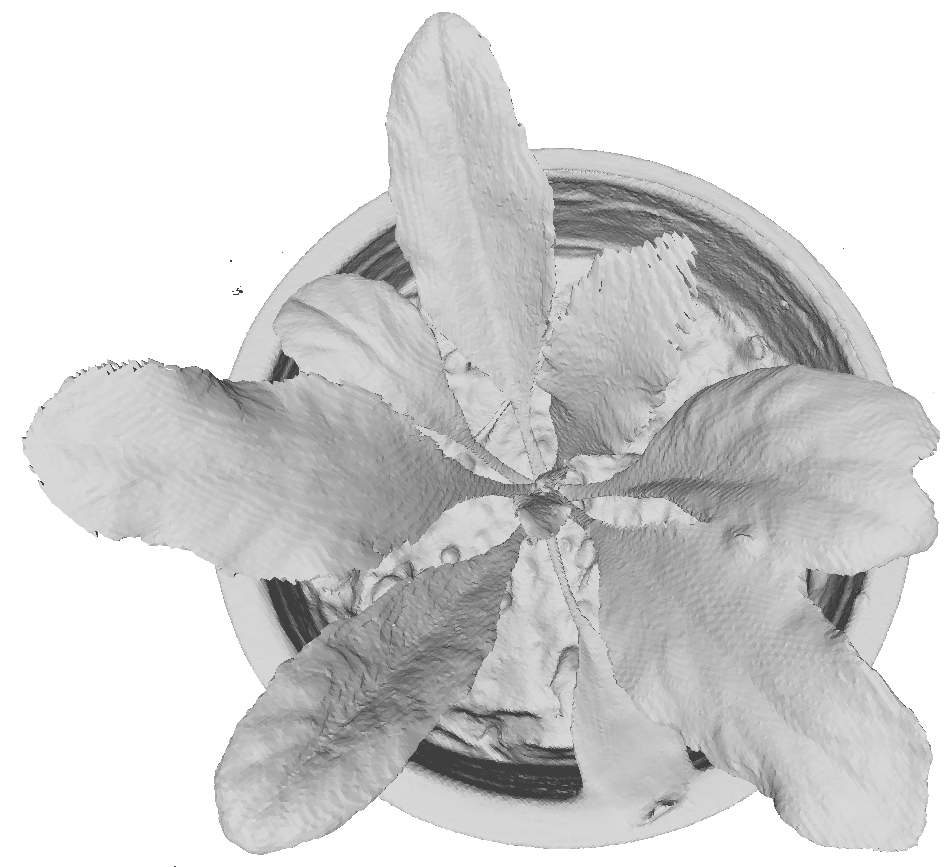}
			\caption{PGSR}
		\end{subfigure}
		\hfill
		\begin{subfigure}[h]{0.23\textwidth}
			\centering
			\includegraphics[width=1\textwidth]{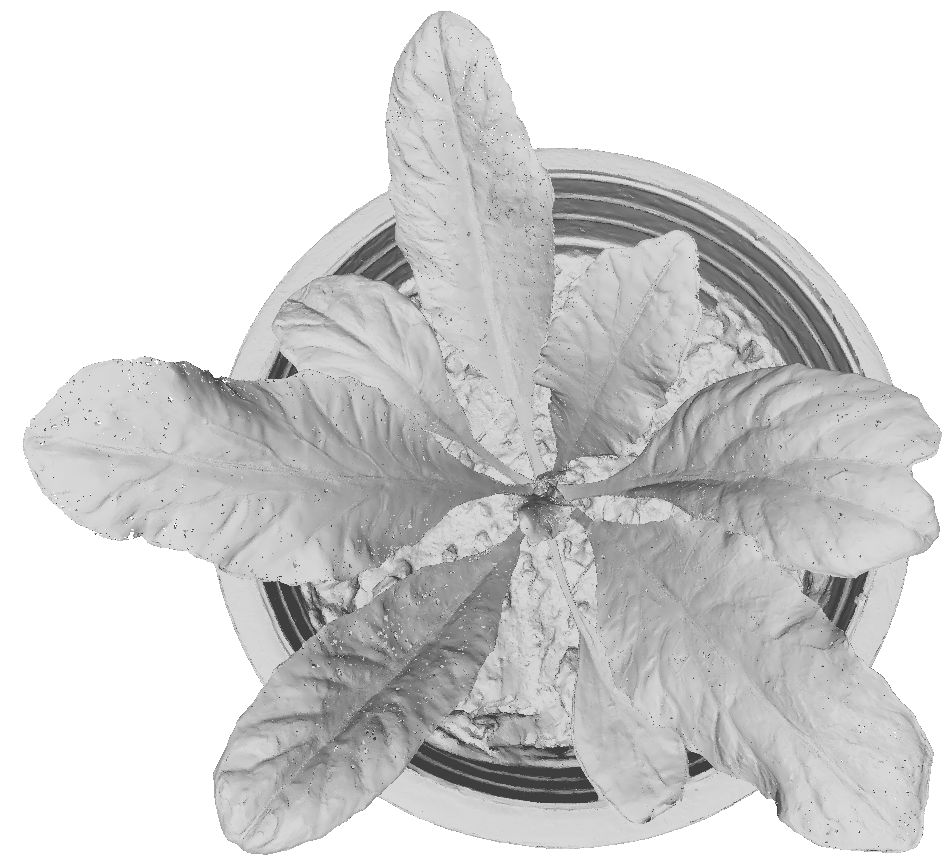}
			\caption{GGGS}
		\end{subfigure}
		\caption{Geometrical Top View of Day 9 Cauliflower for all the Pipelines}
		\label{fig:Grayscale Top View}
	\end{figure}

	The cleaned output meshes, after the removal of floaters from the reconstruction outputs, of the other 6 pipelines except the 3DGS-to-PC can be observed from the front and the top in Figure \ref{fig:Front View} and Figure \ref{fig:Top View}, respectively.
	These figures correspond to the cauliflower plant on the same day (Day 9) output by different pipelines.
	Furthermore, the normal maps corresponding to the same can be seen in Figure \ref{fig:Front View Normals} and Figure \ref{fig:Top View Normals} for the front and top views, respectively.
	Similarly, the mesh geometry obtained from all the pipelines can be seen in Figure \ref{fig:Grayscale Front View} and Figure \ref{fig:Grayscale Top View}.
	
\begin{figure*}[h]
		\begin{subfigure}[h]{1\textwidth}
			\centering
			\includegraphics[width=1\textwidth]{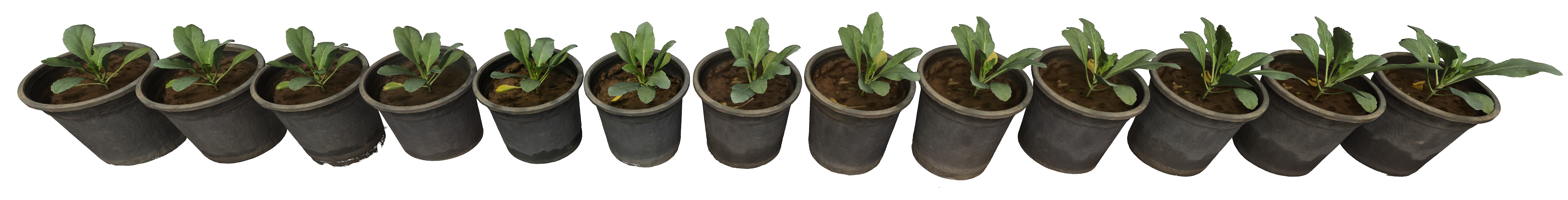}
			\caption{Oblique Frontal View}
		\end{subfigure}
		
		\begin{subfigure}[h]{1\textwidth}
			\centering
			\includegraphics[width=1\textwidth]{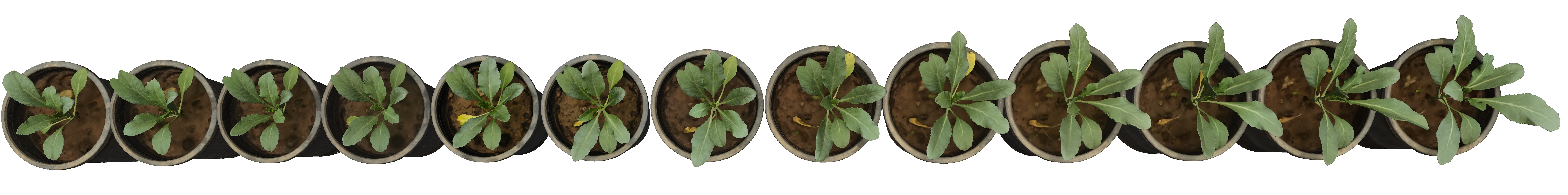}
			\caption{Approximate Top View}
		\end{subfigure}
		\caption{Day 1 to 13 meshes generated via the 2DGS pipeline. These demonstrate the temporal growth of the cauliflower crop.}
		\label{fig:All13plants}
	\end{figure*}

	Although there were very strong ratings for GGGS in almost all aspects, but it had some caveats too, which were mainly its larger mesh size and a large amount of granular artifacts on the mesh surface.
	The 2DGS meshes had a lighter size as compared to the GGGS meshes, which made it possible to load all 13 2DGS meshes at one time in a rendering software like Meshlab~\cite{Cignoni2008meshlab} on a high-end desktop with NVIDIA RTX 4090 32GB VRAM, as can be seen in Figure~\ref {fig:All13plants}.
	This lightweight quality of the meshes is highly preferable to the phenotyping experts, as the simultaneous loading of these multiple meshes is helpful to observe the temporal traits and growth of the crop.

	\subsubsection{Visual Defects}
\begin{figure}
		\centering
		\begin{subfigure}[h]{0.23\textwidth}
			\centering
			\includegraphics[width=1\textwidth]{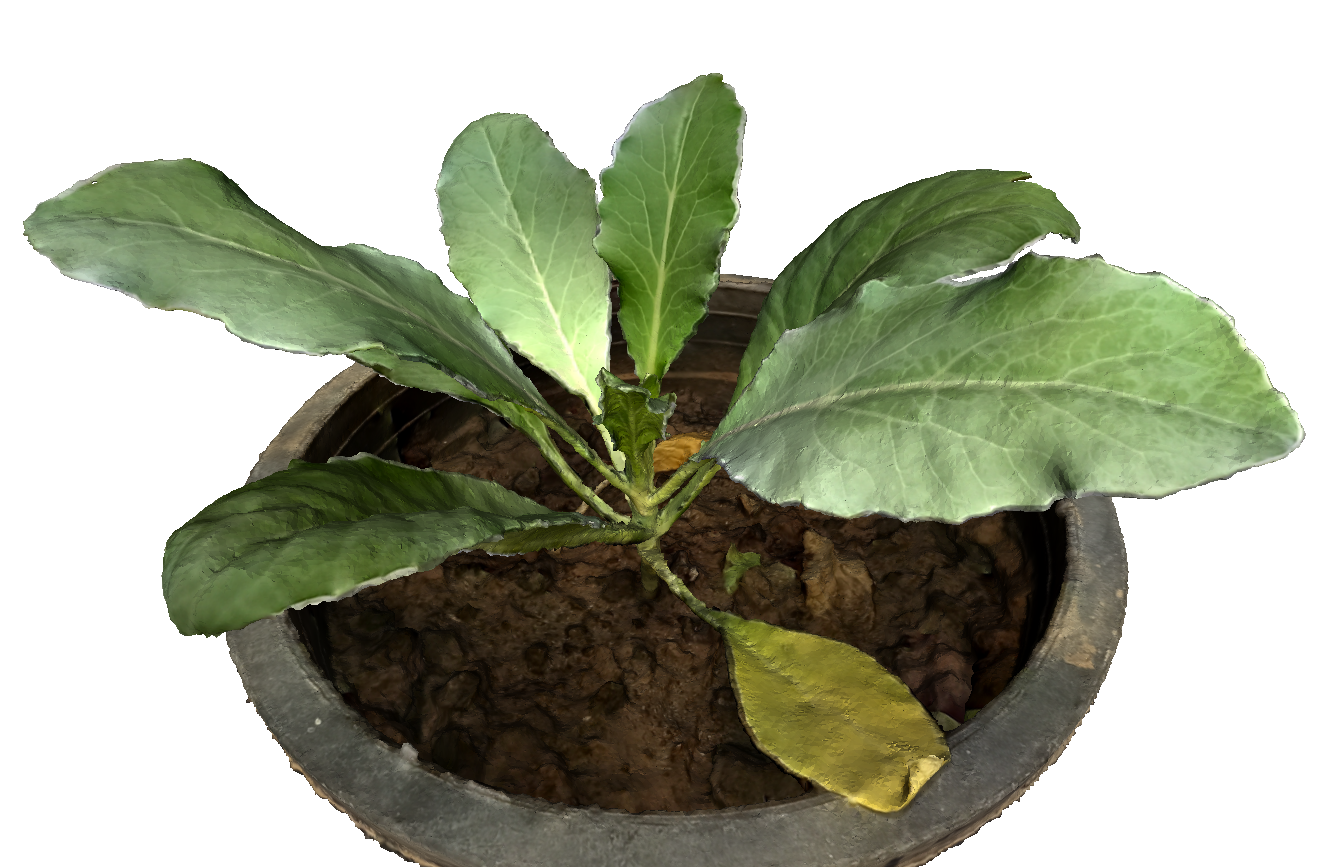}
			\caption{Alicevision Meshroom}
		\end{subfigure}
		\hfill
		\begin{subfigure}[h]{0.23\textwidth}
			\centering
			\includegraphics[width=1\textwidth]{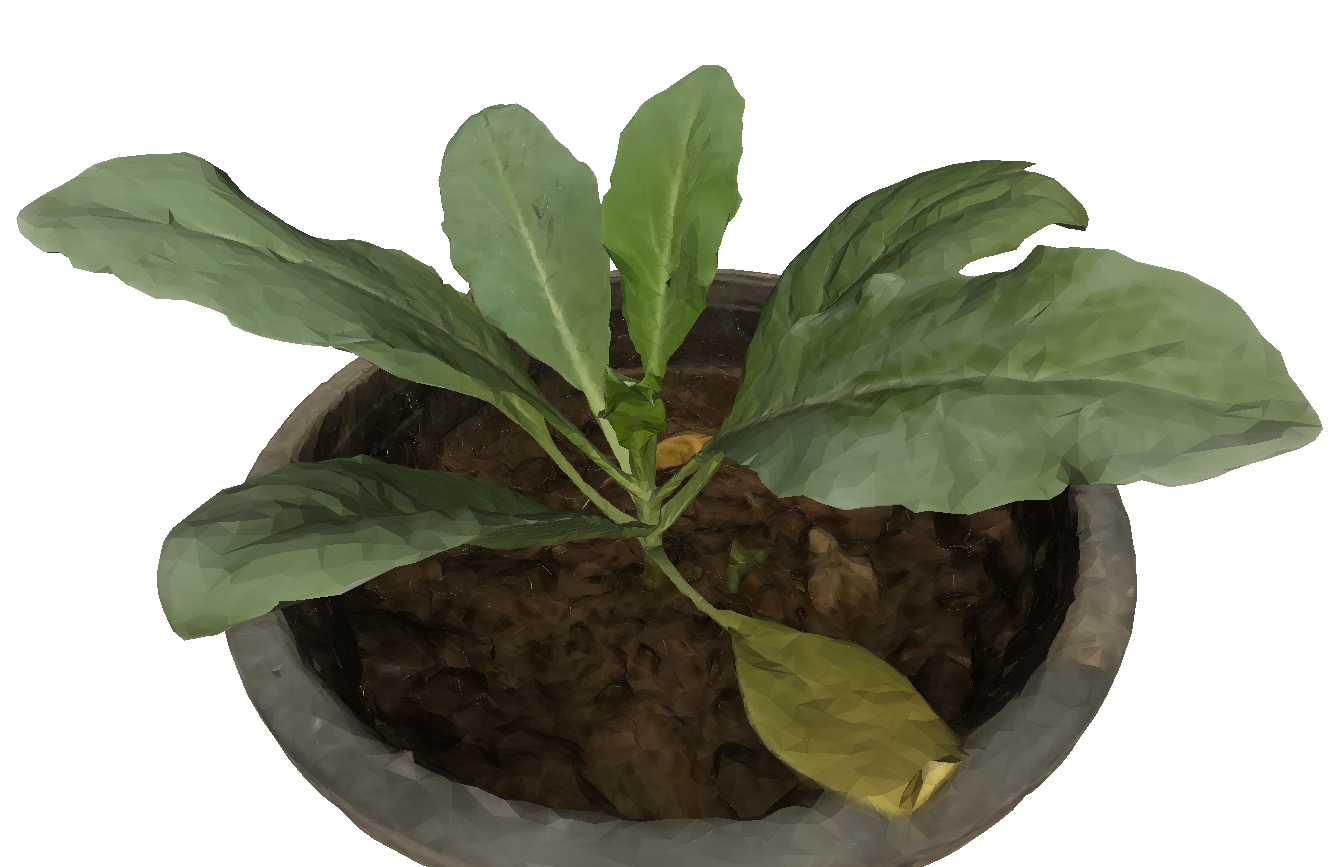}
			\caption{SuGaR}
		\end{subfigure}
		\begin{subfigure}[h]{0.23\textwidth}
			\centering
			\includegraphics[width=1\textwidth]{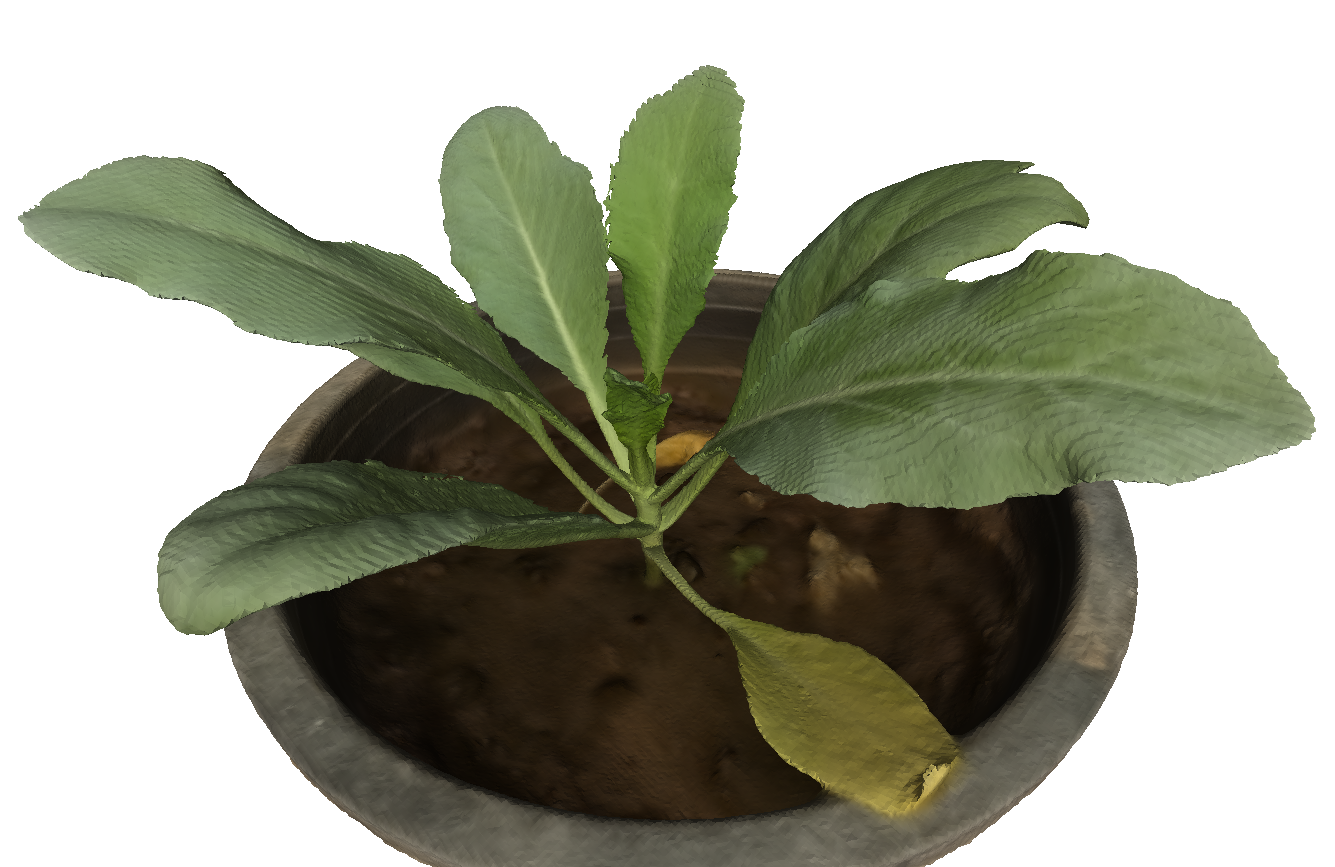}
			\caption{2D Gaussian Splats}
		\end{subfigure}
		\hfill
		\begin{subfigure}[h]{0.23\textwidth}
			\centering
			\includegraphics[width=1\textwidth]{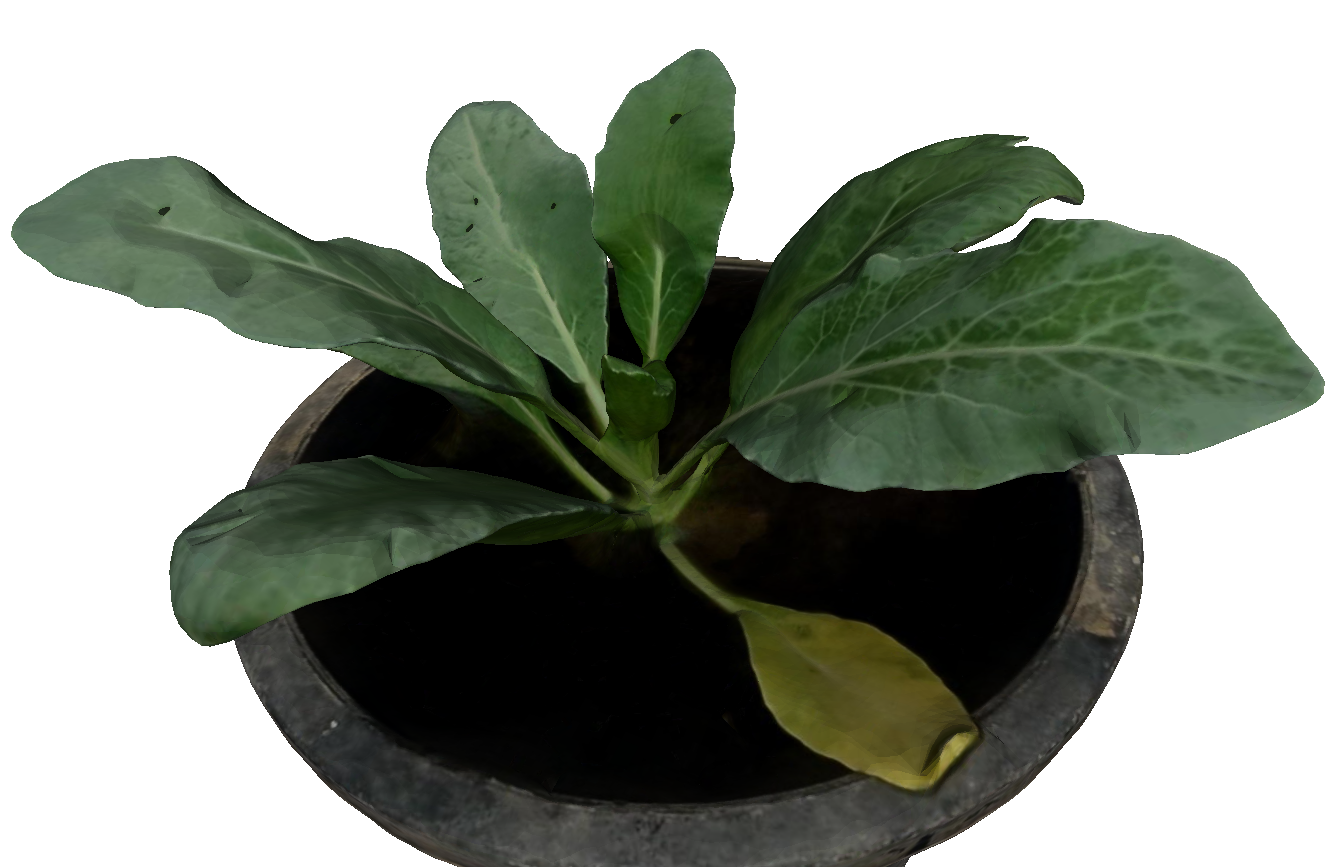}
			\caption{NeRF2Mesh}
		\end{subfigure}		\begin{subfigure}[h]{0.23\textwidth}
			\centering
			\includegraphics[width=1\textwidth]{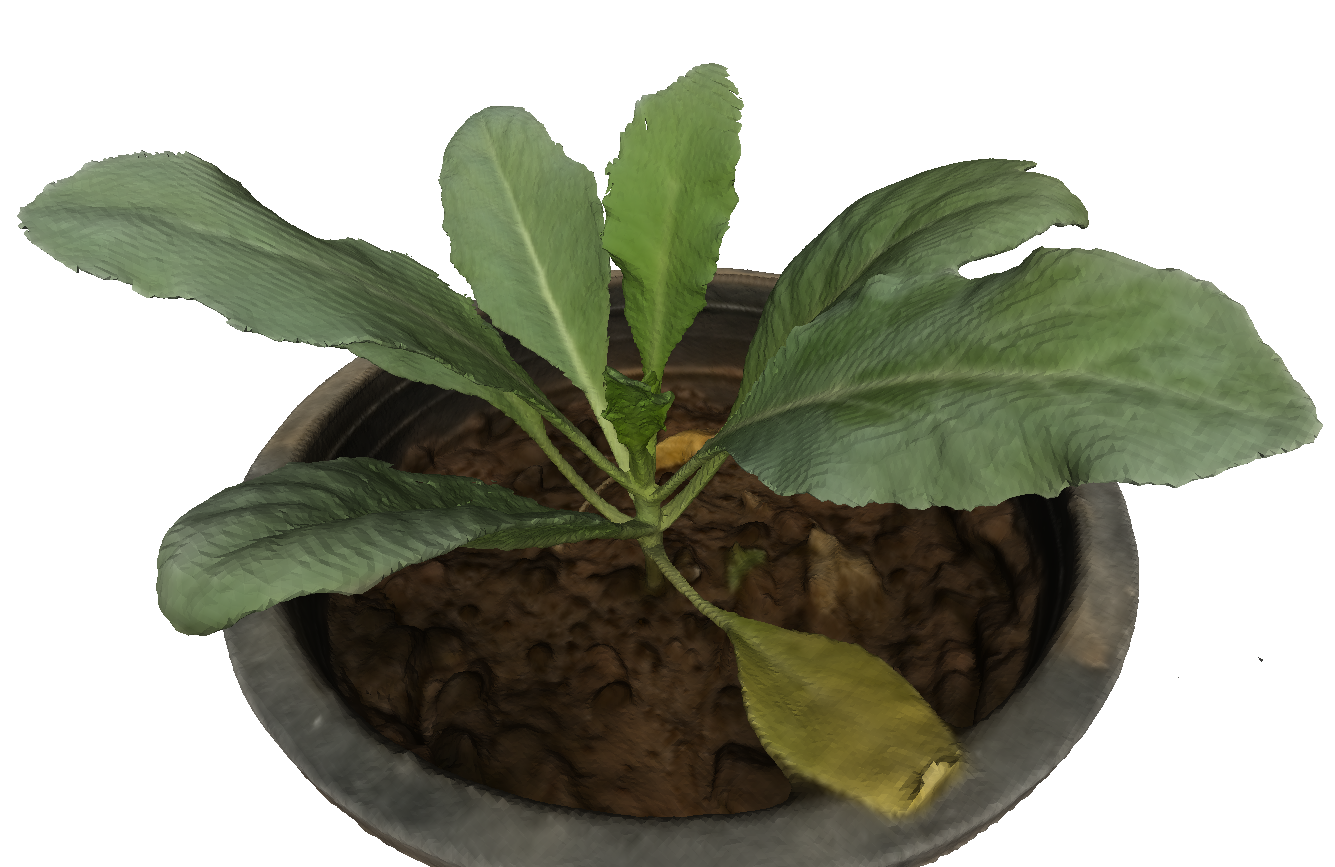}
			\caption{PGSR}
		\end{subfigure}
		\hfill
		\begin{subfigure}[h]{0.23\textwidth}
			\centering
			\includegraphics[width=1\textwidth]{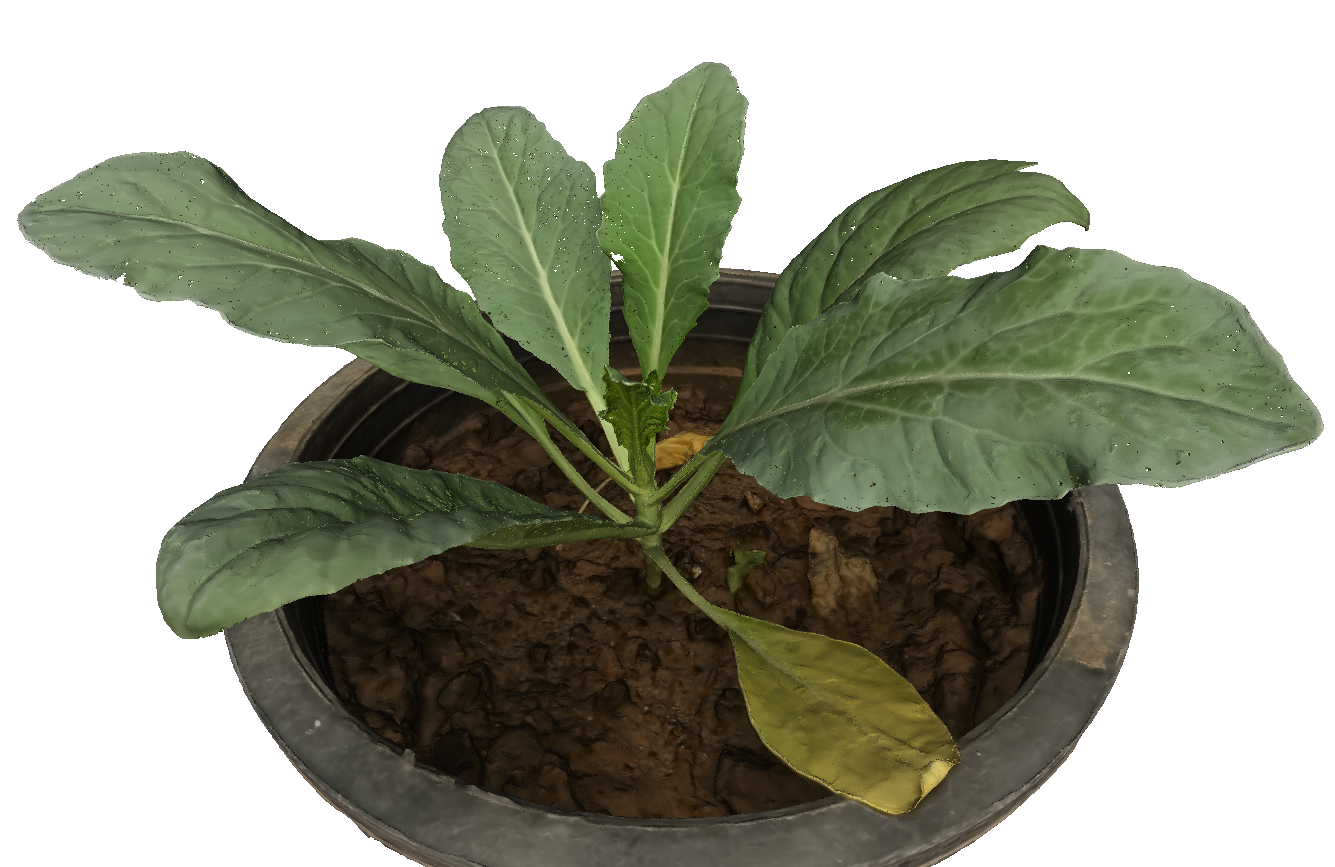}
			\caption{GGGS}
		\end{subfigure}
		\caption{Front View of Day 9 Cauliflower for all the Pipelines}
		\label{fig:Front View}
	\end{figure}

\begin{figure}
		\centering
		\begin{subfigure}[h]{0.23\textwidth}
			\centering
			\includegraphics[width=1\textwidth]{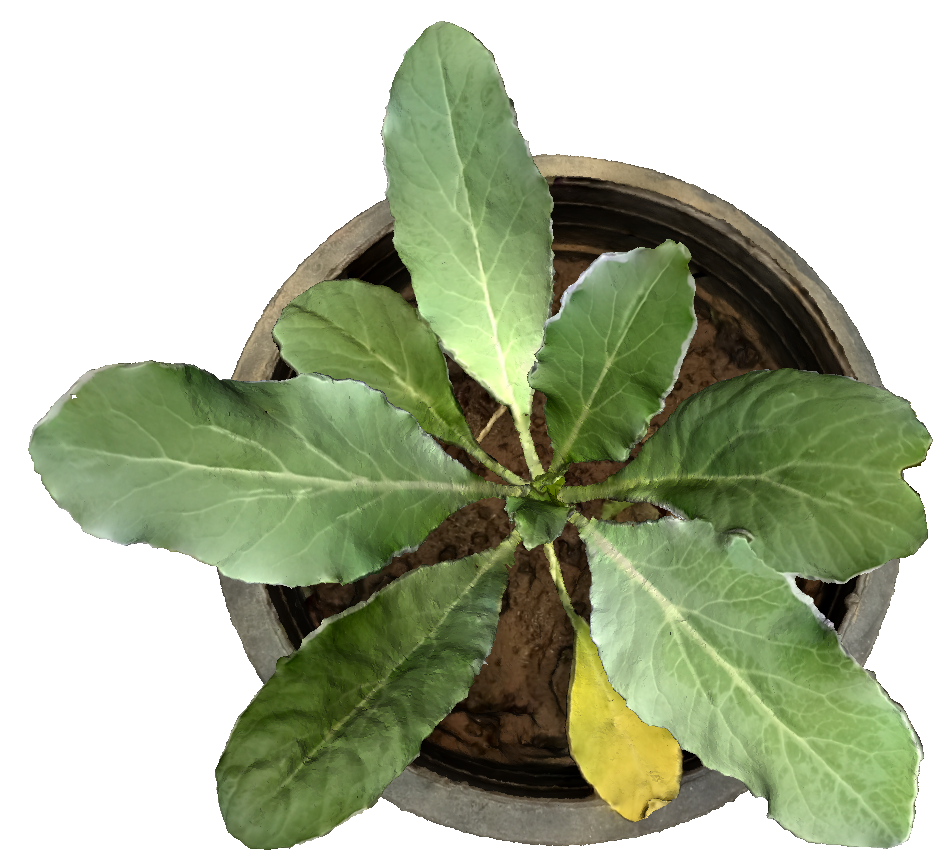}
			\caption{Alicevision Meshroom}
		\end{subfigure}
		\hfill
		\begin{subfigure}[h]{0.23\textwidth}
			\centering
			\includegraphics[width=1\textwidth]{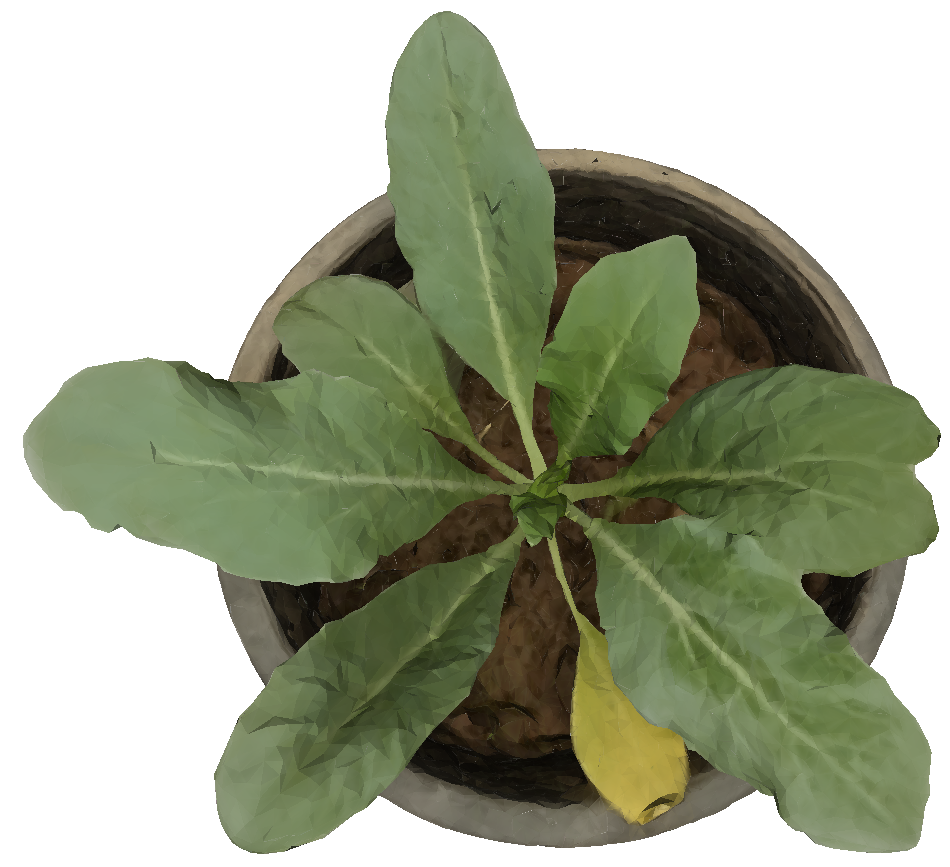}
			\caption{SuGaR}
		\end{subfigure}
		\begin{subfigure}[h]{0.23\textwidth}
			\centering
			\includegraphics[width=1\textwidth]{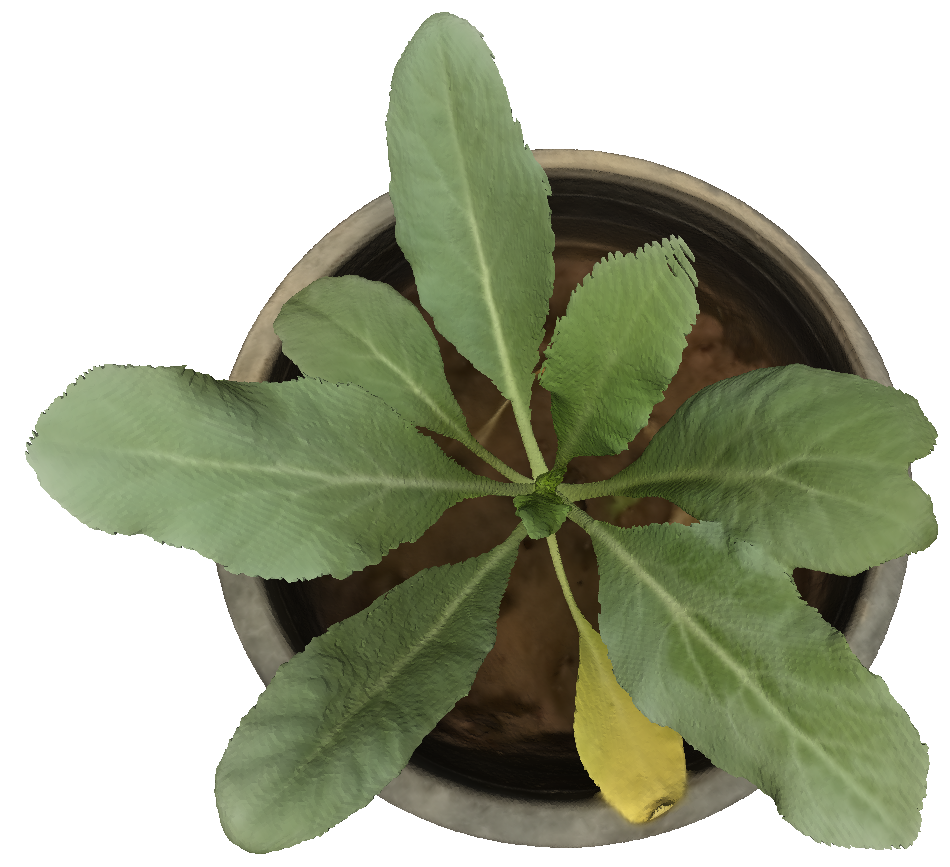}
			\caption{2D Gaussian Splats}
		\end{subfigure}
		\hfill
		\begin{subfigure}[h]{0.23\textwidth}
			\centering
			\includegraphics[width=1\textwidth]{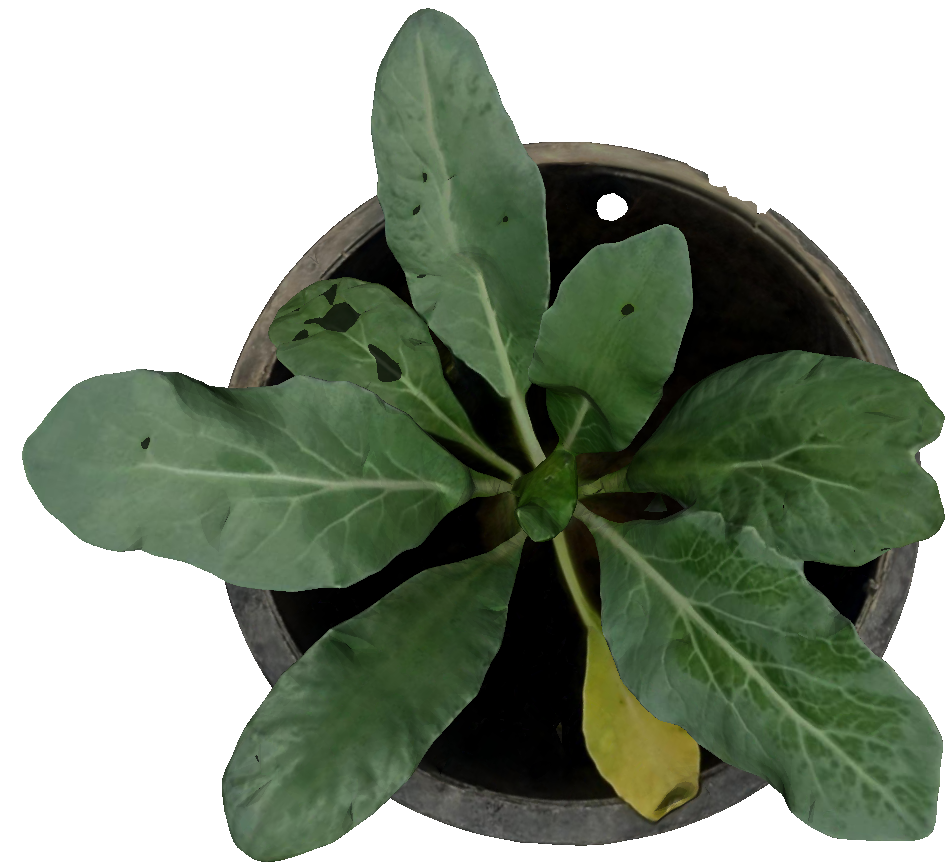}
			\caption{NeRF2Mesh}
		\end{subfigure}
		\begin{subfigure}[h]{0.23\textwidth}
			\centering
			\includegraphics[width=1\textwidth]{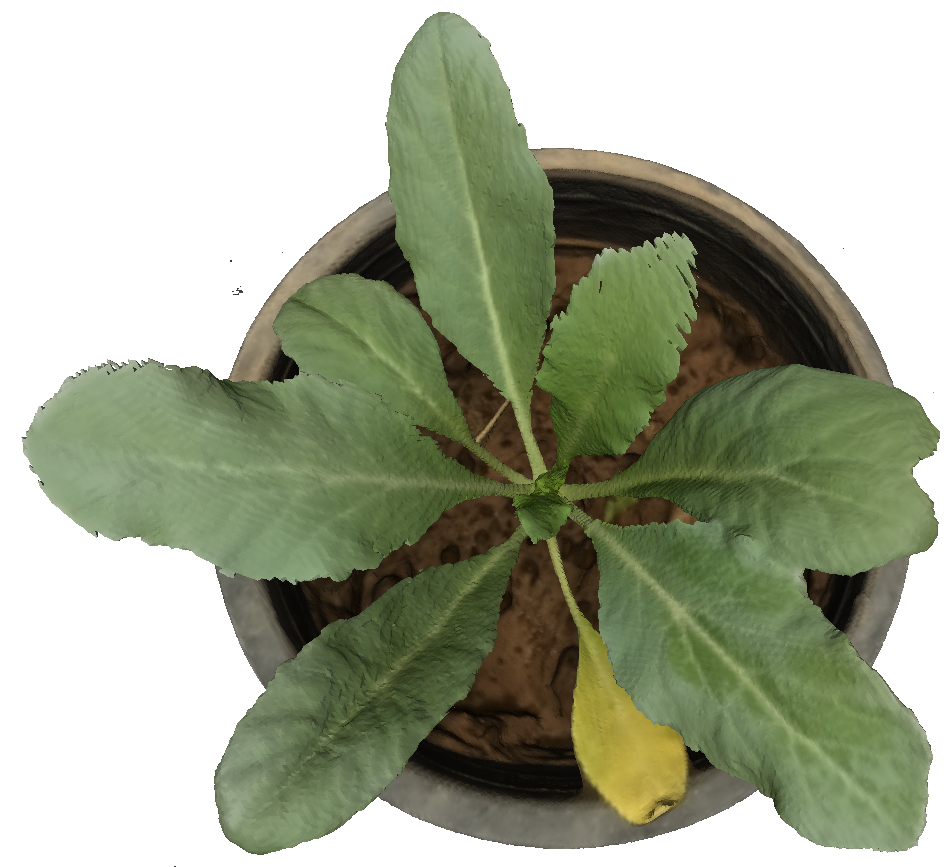}
			\caption{PGSR}
		\end{subfigure}
		\hfill
		\begin{subfigure}[h]{0.23\textwidth}
			\centering
			\includegraphics[width=1\textwidth]{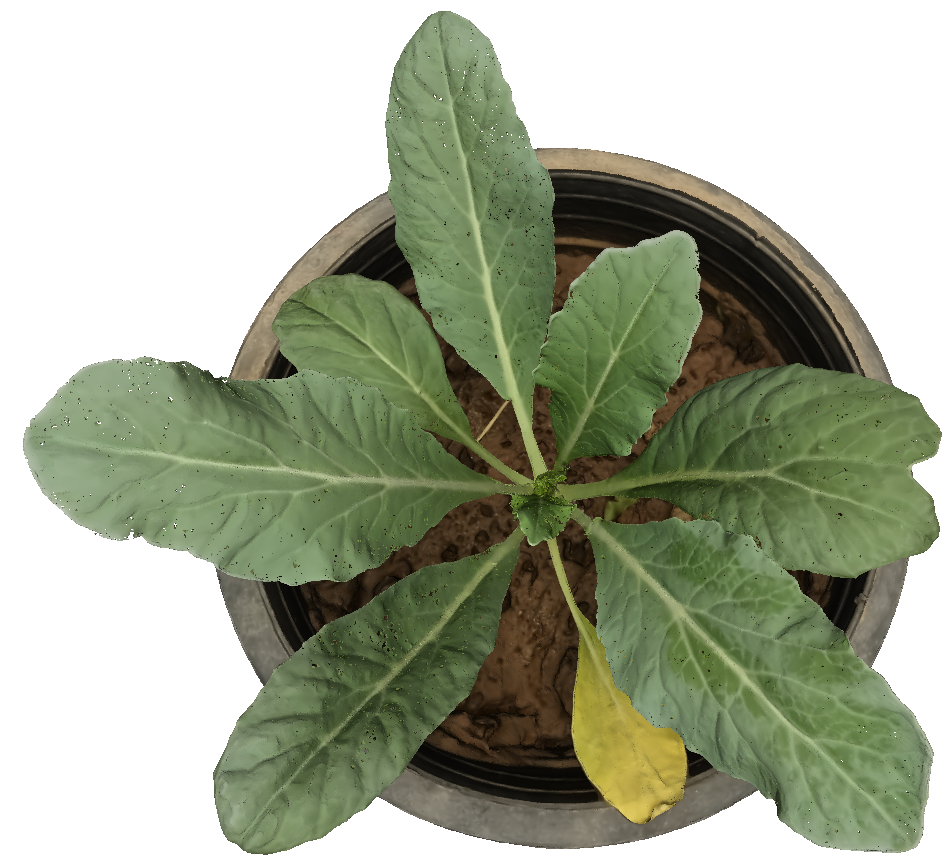}
			\caption{GGGS}
		\end{subfigure}
		\caption{Top View of Day 9 Cauliflower for all the Pipelines}
		\label{fig:Top View}
	\end{figure}

	Considering the visual results, none of the pipelines outputted the perfect mesh like a real plant.
	There were some observable defects with every pipeline briefly mentioned in Table \ref{Approachestable} also. GGGS had multiple tiny granules on the surface of the whole mesh, as can be seen on magnifying Figures \ref{fig:Front View} and \ref{fig:Top View}.
	The mesh surface of the 2DGS and PGSR had contour-like patterns, which are due to the voxel-space-based triangulation procedure involved there.
	Alicevision had issues with reconstruction around the leaf edges.
	NeRF2Mesh had clear visible holes and no visual soil details in the pot region, indicating the lack of reconstruction in the peripheral regions.
	SuGaR's reconstructed surface had a stitched appearance, as various seams were visible when magnifying the surface.
	3DGS-to-PC mesh structure was poor, as can be seen in Figure \ref{fig:pcandmesh}.
		
	\section{Conclusion}
	In this study, we have explored the 3D reconstruction prowess of seven pipelines, namely Alicevision Meshroom, 3DGS-to-PC, SuGaR, 2DGS, NeRF2Mesh, PGSR, and GGGS.
    We have used our unreleased \emph{Cauliflower-13} dataset as the pipeline inputs.
    We evaluated them based on 4 quantitative metrics, namely, Chamfer distance, PSNR, LPIPS, and SSIM.
    Alongside, we also conducted a user study with a questionnaire curated to the phenotyping aspects of the crops.
    Experiments on our dataset revealed that recent Gaussian Splatting-based methods have outperformed NeRF-based and traditional photogrammetry-based methods.
	In particular, the GGGS pipeline is the most preferred, and it beats the second-best pipeline (2DGS) by around 27\% on the radar chart.
    However, 2DGS is more efficient in terms of storage space, which is also an important factor to consider for temporal phenotyping.
    Overall, all the considered pipelines had some notable defects, implying there is scope for the perfect 3D reconstruction pipeline.
	But on an ending note, the ongoing advancements in the field of Gaussian Splatting are still promising as they bridge the gap between the traditional graphics pipelines and the recent rendering-based paradigms, making it beneficial from a practical standpoint of speed and efficiency.

\section*{Acknowledgements}
    The work was supported by TIH-AwaDH at IIT Ropar, which is a technology innovation hub for agriculture \& water technology development sponsored by DST.
    
	\bibliographystyle{cas-model2-names}
	
	\bibliography{cas-refs}

	
	
	
	
\end{document}